\documentclass[letterpaper]{article} 
\usepackage{aaai2027}  
\usepackage[hyphens]{url}  
\usepackage{graphicx} 
\usepackage{natbib}  
\usepackage{caption} 
\usepackage{algorithm}
\usepackage{algorithmic}
\usepackage{xcolor}
\usepackage{amsmath}

\newif\ifreviewmode
\reviewmodetrue               

\ifreviewmode
  \definecolor{revcol}{RGB}{0,80,190}
  \long\def\del#1{\textcolor{gray}{\sout{#1}}}     
  \newenvironment{revblock}{\color{revcol}}{}      
\else

  \long\def\del#1{}
  
\fi

\usepackage{newfloat}
\usepackage{listings}
\DeclareCaptionStyle{ruled}{labelfont=normalfont,labelsep=colon,strut=off} 
\floatstyle{ruled}
\newfloat{listing}{tb}{lst}{}
\floatname{listing}{Listing}

\usepackage{booktabs}

\usepackage{multirow}
\usepackage{xspace}
\newcommand{\newtask}{SimuScene\xspace}

\newcommand{\Nmodels}{10\xspace}

\newcommand{\Nconcepts}{52\xspace}

\newcommand{\figref}[1]{Figure~\ref{#1}}
\newcommand{\tabref}[1]{Table~\ref{#1}}

\title{Training and Benchmarking Code Generation for Physics-Inspired Animations}
\author {
    Yanan Wang\textsuperscript{\rm 1}\equalcontrib,
    Renxi Wang\textsuperscript{\rm 1}\equalcontrib,
    Yongxin Wang\textsuperscript{\rm 1}\equalcontrib,
    Xuezhi Liang\textsuperscript{\rm 1},
    Fajri Koto\textsuperscript{\rm 1},
    Timothy Baldwin\textsuperscript{\rm 1},
    Xiaodan Liang\textsuperscript{\rm 1,\rm 2}\corresponding,
    Haonan Li\textsuperscript{\rm 1}
}
\affiliations {
    \textsuperscript{\rm 1}Mohamed bin Zayed University of Artificial Intelligence\\
    \textsuperscript{\rm 2}Sun Yat-sen University\\
    yanan.wang@mbzuai.ac.ae, renxi.wang@mbzuai.ac.ae, yongxin.wang@mbzuai.ac.ae, xuezhi.liang@mbzuai.ac.ae, fajri.koto@mbzuai.ac.ae, timothy.baldwin@mbzuai.ac.ae, xiaodan.liang@mbzuai.ac.ae, haonan.li@mbzuai.ac.ae
}

\begin{document}

\maketitle

\begin{abstract}
Large language models (LLMs) have been widely studied in areas such as mathematical reasoning, complex coding, and scientific problem solving. However, their ability to generate executable code that visually depicts physical scenarios and their qualitative dynamics remains underexplored. We propose SimuScene, the first systematic study that trains and evaluates LLMs on code generation for physics-inspired animations across 52 concepts spanning five physics domains. We build an automated data collection pipeline with human verification to ensure data quality. The resulting dataset contains 7,659 scenarios, including a 334-example human-verified test set. We evaluate 10 contemporary LLMs and find that even the strongest model achieves only a 21.5\% Avg@8 accuracy, demonstrating the difficulty of generating animations that are both executable and visually aligned with physical scenario descriptions. Finally, we introduce a reinforcement learning pipeline that uses visual rewards from code-generated videos to train text-only LLMs, with a vision-language model evaluating videos through verification questions. Experiments show that training with our data and video-based rewards improves LLM performance on physics-inspired animation generation.
\end{abstract}

\section{Introduction}

Modern large language models (LLMs) \citep{openai_o3_o4_mini_2025,deepseekai2025deepseekr1incentivizingreasoningcapability,qwen3technicalreport,cheng2025k2thinkparameterefficientreasoning} exhibit strong reasoning across domains, including mathematics \citep{maaInvitationalCompetitions}, coding \citep{evalplus,jain2024livecodebench}, science \citep{zhang2025physreason}, and agent tasks \citep{luo2025mcp,merrill2026terminal}. However, using code to generate physics-inspired animations that visually depict physical scenarios and qualitative dynamics remains underexplored. Existing benchmarks evaluate symbolic reasoning, coding, or perceptual understanding, but do not require models to generate executable code whose rendered animations are visually aligned with dynamic physical scenario descriptions.

Physics-inspired animation generation is inherently tied to motion, temporal evolution, and object interactions. Producing such animations requires both qualitative physical understanding and the coding ability to represent objects, trajectories, and interactions in executable visual form. This makes it a suitable task for evaluating LLMs at the intersection of physical-scenario understanding and animation-oriented code generation. Rather than targeting numerically precise simulation, we focus on whether LLMs can generate executable code that produces visually interpretable animations consistent with qualitative physical descriptions.

In this paper, we study the task of \textbf{physics-inspired animation generation}, and introduce \textbf{SimuScene}, a novel task, dataset, and benchmark for evaluating this capability. The task emphasizes rendered animations that preserve basic physical plausibility, capture qualitative dynamics, and visually match the input description. This focus on functional and perceptual consistency supports downstream applications such as educational visualization, controllable animation prototyping, and visual reasoning. Throughout the paper, \newtask refers to this proposed task, dataset, and benchmark.

\begin{figure*}[t]
  \centering
  \includegraphics[width=.9\textwidth]{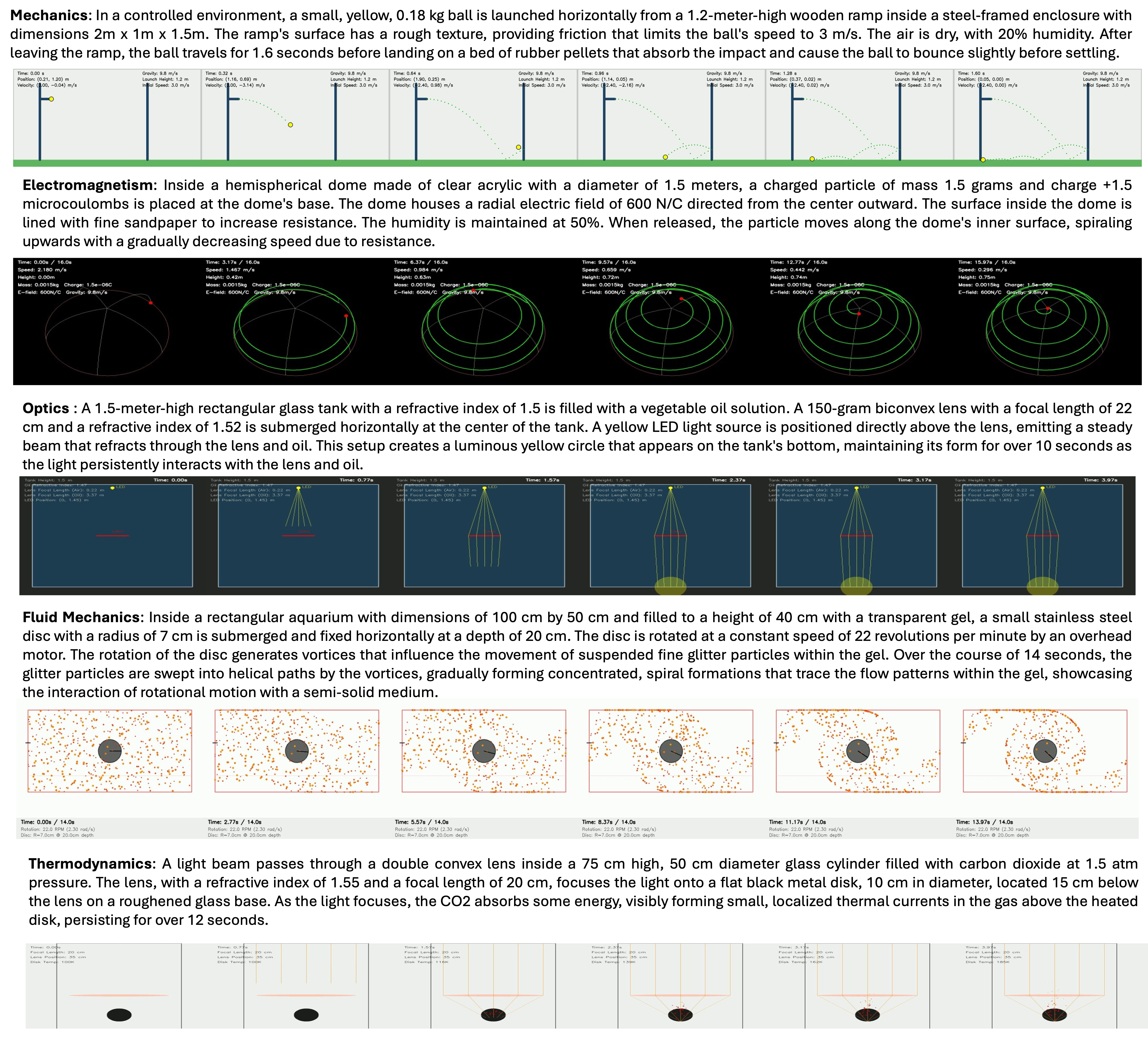}
  \caption{Representative text-to-code-to-video examples from \newtask. Each video is rendered from LLM-generated code and depicts the qualitative dynamics described in the corresponding physical scenario.}
  \label{fig:data_example}
\end{figure*}

To construct \newtask, we propose an automatic pipeline that starts from \Nconcepts concepts across five physics domains and generates difficulty-controlled physical scenarios. Each scenario is paired with verification questions describing expected qualitative dynamics, and the data are checked through automated validation; see ``SimuScene'' for details. The dataset contains 7,659 scenarios across mechanics, electromagnetism, optics, fluid mechanics, and thermodynamics. Each centers on a single core concept, with one to three additional conditions controlling difficulty. We curate a 334-example human-verified test set, with at least four verification questions per scenario; the remaining scenarios are used for training. Manual inspection of 100 sampled training examples shows an average quality rate of 84\%, supporting the reliability of the automatic pipeline. Representative examples and code-generated animations are shown in \figref{fig:data_example}.

We evaluate \Nmodels state-of-the-art LLMs on \newtask and find that the best model achieves only 21.5\% Avg@8 accuracy. These results indicate that existing LLMs struggle to generate executable code whose rendered animations align with textual descriptions and capture plausible qualitative dynamics, highlighting \newtask as a challenging benchmark for scene-level understanding and code generation.

Finally, we explore whether training improves performance on \newtask. We introduce a \textbf{Code-Video-Judge} reward signal from code-generated videos, with design choices that provide stable, fine-grained feedback during reinforcement learning (RL), and validate the pipeline through human evaluation. Using this reward, our trained 7B model achieves an 11.1\% pass rate, compared to 0\% for the baseline, showing the effectiveness of our dataset and training method. Moreover, mixing our data with an instruction-following dataset (see the supplementary document) yields substantial gains in coding performance. To our knowledge, this work is among the first to use vision-based rewards to train text-only models.

To summarize, our work makes three contributions centered on a new evaluation paradigm for code-generated physics-inspired animations:
\begin{enumerate}
    \item We present \textbf{\newtask}, a novel task, dataset, and benchmark for evaluating LLMs on physical-scenario understanding, qualitative dynamics, and animation-oriented code generation, with a rigorously human-verified test set of 334 scenarios and quality-controlled training data.
    \item We evaluate \Nmodels frontier LLMs, showing that even the strongest achieves only a $\sim$21\% Avg@8 accuracy, highlighting model limitations and the challenge of \newtask.
    \item We further explore reinforcement learning with signals from code-generated videos, yielding substantial gains in physics-inspired animation generation for smaller models. We also validate the reliability of VLM-based judging and find no evidence of reward hacking.
\end{enumerate}

\section{Related Work}
\textbf{Reasoning Evaluation for LLMs.}
Recent work improves LLM reasoning by equipping models with complex thinking before generating their response. Such LLMs include OpenAI's o-series \citep{openai_o1_api_docs,openai_o3_o4_mini_2025}, DeepSeek-R1, V3.1 \citep{deepseekai2025deepseekr1incentivizingreasoningcapability,deepseekai2024deepseekv3technicalreport}, K2-Think \citep{cheng2025k2thinkparameterefficientreasoning}, and Qwen3~\citep{qwen3technicalreport}. As these capabilities advance, benchmarking becomes vital. Recent work has focused on evaluating LLMs' reasoning capabilities along several dimensions. {On coding, the SWE-Bench~\citep{jimenez2024swebench,openai2024swebenchverified,deng2025swe} series has been widely used for benchmarking LLMs on fixing repository-level bugs. CodeForces-style benchmarks~\citep{quan2025codeelo} focus on competition-level programming problems. For mathematics, AIME-26~\citep{maaInvitationalCompetitions} and IMO-Bench~\citep{luong2025towards} provide difficult math problems from real competitions. FrontierMath~\citep{glazer2024frontiermath} evaluates LLMs on most major branches of modern mathematics. For natural science, GPQA-Diamond~\citep{rein2023gpqa} comprises expert-written multiple-choice questions in biology, physics, and chemistry. Humanity's Last Exam~\citep{phan2025humanity} tests expert-level questions across many disciplines and remains far from saturated. \newtask, on the other hand, measures the underexplored LLM capability of generating executable code whose rendered animations visually depict qualitative physical dynamics.

\textbf{Physical Understanding and Reasoning with LLMs.}
Physical understanding requires spatial relationships, temporal dynamics, and motion patterns. Existing benchmarks evaluate this through static plausibility judgments and scenario-based QA~\citep{zellers2018swaglargescaleadversarialdataset,bisk2019piqareasoningphysicalcommonsense}, physics problem solving~\citep{wang2023newtonlargelanguagemodels,xu2025ugphysicscomprehensivebenchmarkundergraduate}, or broader physical reasoning over predefined problems~\citep{qiu2025phybenchholisticevaluationphysical,zhang2025abenchphysicsbenchmarkingphysicalreasoning}. Other approaches integrate simulators for analysis-by-synthesis reasoning~\citep{cherian2024llmphycomplexphysicalreasoning}. In contrast, \newtask evaluates a complementary generative capability: translating natural-language scenario descriptions into executable code that produces visually aligned animations capturing qualitative dynamics. A detailed comparison with physics reasoning and simulation is provided in the supplementary document.

\textbf{LLM/VLM-as-Judge Evaluation.} Model-based evaluation has become standard where outputs are open-ended and no unique reference exists. \citet{zheng2023judging} show that GPT-4 judges reach over 80\% agreement with human preferences, which is the same level humans reach with each other. HLE \citep{phan2025humanity} relies on an LLM judge to verify answer equivalence.
VIEScore \citep{ku2024viescore} extends VLM judging to conditional image synthesis, reaching correlation with human ratings approaching the human-to-human ceiling. The same paradigm underlies default leaderboards in image generation \citep{hu2023tifa,cho2024davidsonian}, image editing \citep{liu2025step1x}, video generation \citep{he2024videoscore}, and extends to agentic evaluation, where WebVoyager \citep{he2024webvoyager} reaches 85.3\% agreement with humans, again matching inter-annotator agreement.
SimuScene extends it in two respects: our artifacts are \emph{temporal}, requiring judgments over motion rather than static layout, and our judge is execution-grounded.

\textbf{RL from Vision Signal.} RL with vision signals has been explored in visual reasoning~\citep{wang2024eaco},  robotics~\citep{zitkovich2023rt}, visual navigation~\citep{wang2019reinforced}, and embodied decision-making~\citep{huang2025vlm}, primarily targeting vision-centric tasks. 
In LLM training, most prior work relies on verifiable ground-truth answers or an LLM to judge response quality \citep{cheng2025revisitingreinforcementlearningllm}. However, for our setting, judging responses alone is insufficient to capture the qualitative dynamics in physical scenarios, since the final behavior is only observable after executing the code and inspecting the rendered video. We introduce a \textbf{Code-Video-Judge} RL paradigm, where rewards are computed by aggregating scores from multiple verification questions answered by VLMs over code-generated videos. Our approach evaluates not only textual reasoning accuracy but also whether the rendered animation captures the qualitative dynamics specified by the input scenario.


\section{\newtask}
This section presents the construction, quality analysis, and evaluation pipeline of the \textbf{\newtask} dataset. Rather than relying directly on raw LLM outputs, we design a multi-stage pipeline that combines LLM-based plausibility filtering, verification-question validation, and human validation to enable scalable and reliable data generation.

\begin{figure*}[t]
  \centering
  \includegraphics[width=.85\textwidth]{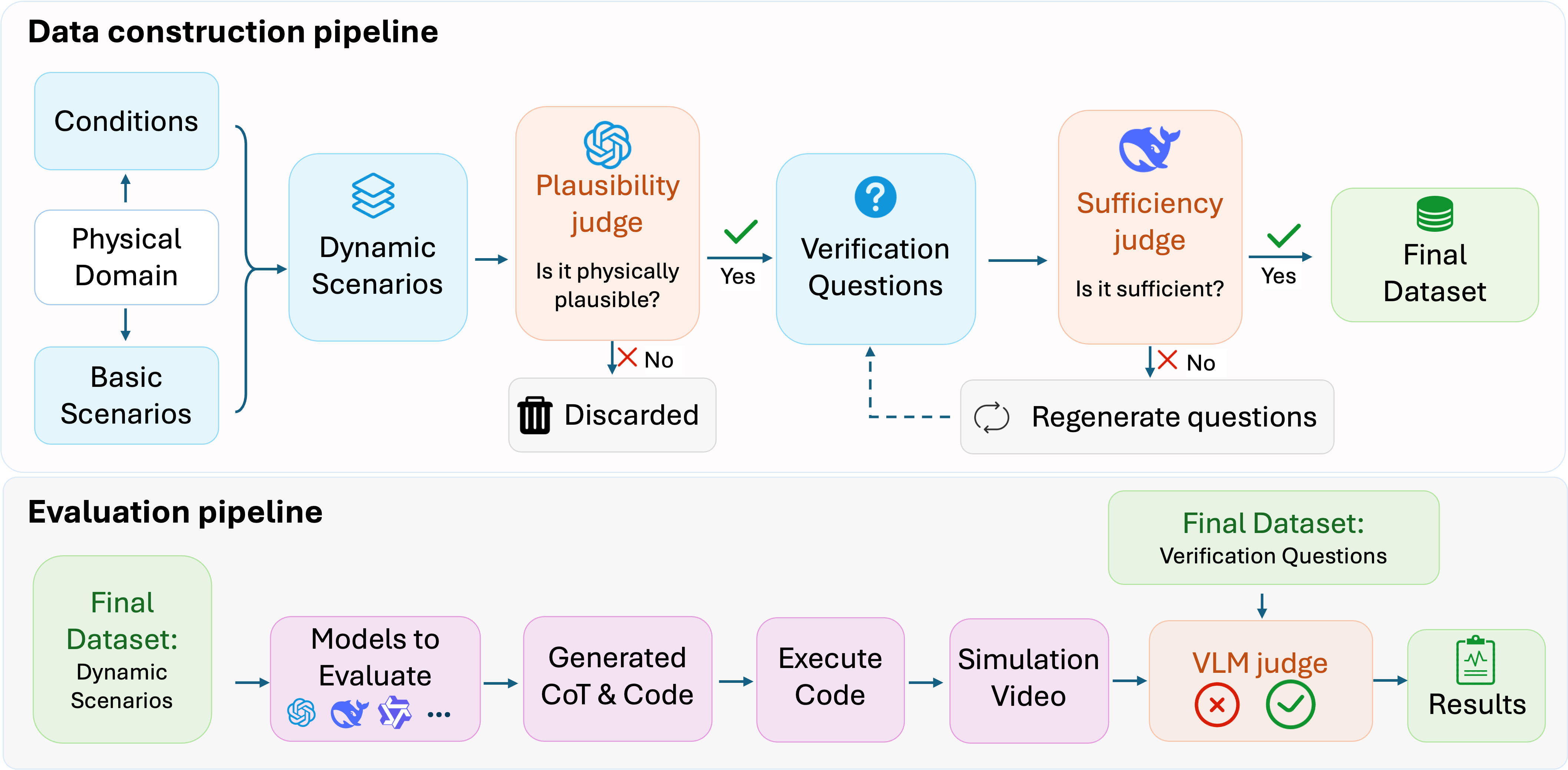}
  \caption{Overview of the \newtask dataset construction pipeline and evaluation pipeline.}
  \label{fig:data_pipeline}
\end{figure*}

\subsection{Dataset Construction}
\textbf{Domain Selection.} We construct a dataset covering diverse \emph{visualizable} dynamic concepts. We select five classical physics domains: mechanics, electromagnetism, optics, fluid mechanics, and thermodynamics, due to their observable motion patterns and qualitative dynamics suitable for code-generated animations. Using a two-level taxonomy aligned with standard physics categorizations \citep{karaoglu2020classical}, we curate 52 concepts (e.g., fluid dynamics, angular momentum, and gravity). See the supplementary document for details.

\textbf{Scenario \& Verification Question Generation.} The construction process is shown in~\figref{fig:data_pipeline}. Starting from 52 concepts, we use GPT-4o~\citep{openai2024gpt4ocard} to generate \textit{basic physical scenarios} (e.g., ``\texttt{A spinning disk on a frictionless table}'') and \textit{additional conditions} (e.g., ``\texttt{Considering air resistance}''). We combine them into dynamic scenarios, with the number of conditions controlling difficulty. Each candidate is automatically checked by GPT-4o for physical plausibility. For valid scenarios, GPT-4o generates \emph{visual verification questions} describing the expected qualitative dynamics. These questions are framed over the rendered video, and all must be answered as \textsc{true} to indicate a visually consistent animation.

\textbf{Automatic Question Validation.} To improve data quality, we introduce a second review stage for verification questions. We use DeepSeek-R1-0528~\citep{deepseekai2025deepseekr1incentivizingreasoningcapability} to assess whether each question set is sufficient to judge the full motion described by its scenario, assuming all answers are ``Yes.'' The reviewer summarizes the key entities and relationships in the scenario, and determines whether the questions adequately cover the expected qualitative dynamics. Insufficient question sets are revised or regenerated to improve coverage. This multi-stage pipeline combines scenario plausibility filtering with independent question review to improve data integrity before human validation. \figref{fig:single_example} shows an example scenario with its verification questions.

\textbf{Dataset Statistics \& Splits.} 
Following this pipeline, we construct 7,659 validated scenarios with verification questions: 7,325 for training and 334 for evaluation. The test set contains at least four verification questions per concept to ensure coverage and reduce domain and concept bias.

For training, we generate eight responses per scenario using DeepSeek-R1-0528~\citep{deepseekai2025deepseekr1incentivizingreasoningcapability}. Our evaluation pipeline labels each response for correctness, yielding 12,556 correct responses spanning 4,182 scenarios whose animations pass the visual verification questions. The remaining 3,143 scenarios without correct responses are used for RL.

To reduce RL difficulty and provide a warm start, we allocate 3,000 scenarios with verified correct responses for SFT. The remaining 1,182 scenarios are combined with 3,143 scenarios without correct responses to form 4,325 RL training scenarios. This split lets the model learn patterns via SFT before tackling both easier and harder scenarios during RL.

Additional statistics, including domain distribution, scenario length distribution, and verification question counts, are provided in the supplementary document.

\begin{figure*}[t]
  \centering
  \includegraphics[width=.9\textwidth]{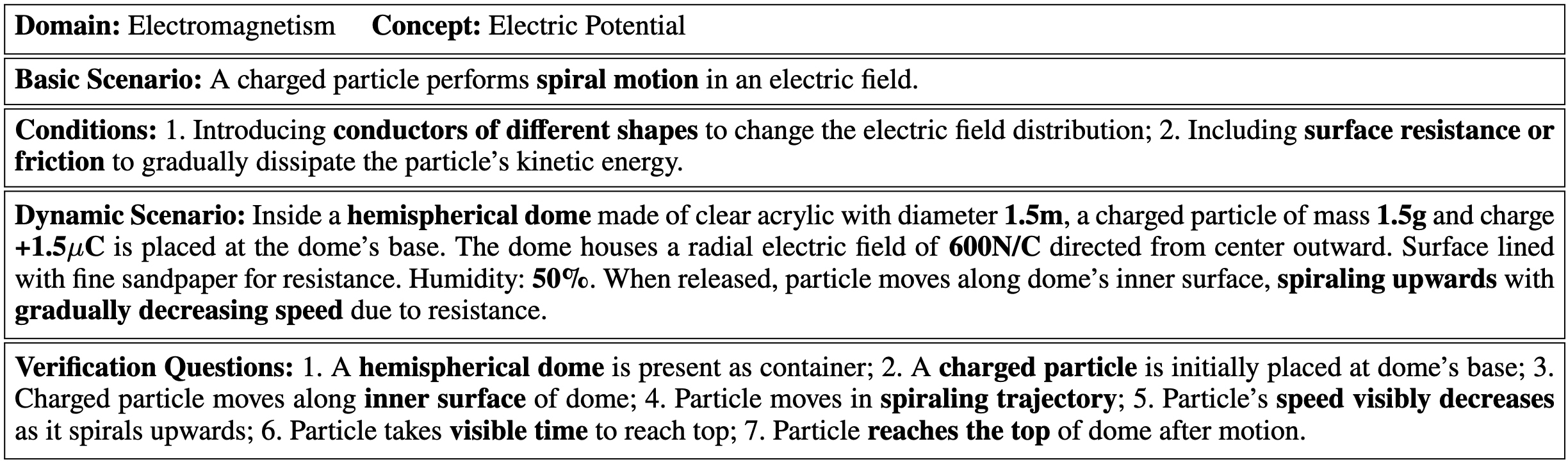}
  \caption{An example data entry from \newtask, including a physical scenario and its associated verification questions. The corresponding animation is shown as the second video example in \figref{fig:data_example}.}
  \label{fig:single_example}
\end{figure*}

\subsection{Dataset Quality Control}
\textbf{Human Verification of Test Set.} 
To ensure test set quality, we manually verified all 334 test scenarios and corrected problematic cases when necessary. Each scenario was reviewed to confirm that: (1) the description is clear and unambiguous; (2) the scenario is plausible and renderable as a code-generated animation; and (3) the verification questions cover key aspects of the expected qualitative dynamics. Ambiguous descriptions, implausible scenarios, or incomplete verification questions were revised or removed. This process ensures that the test set serves as a reliable benchmark for evaluating LLM-generated physics-inspired animations.

\textbf{Training Data Quality Assessment.} 
To assess training data quality, we manually inspected 100 random examples: 50 from the SFT data and 50 from the RL data, using the same criteria as above. The results show that 88.5\% of SFT samples and 80.0\% of RL samples meet our quality standards, indicating the robustness of our automated data construction pipeline. Most issues arise from ambiguous physical descriptions and incomplete verification coverage. The lower quality rate for RL data is expected: SFT data underwent VLM-based filtering to retain only visually verified correct examples, while RL data includes scenarios with both successful and failed generations to provide diverse training signals.

\subsection{Evaluation Pipeline}\label{sec:eval_protocol}

\textbf{Code–Video–VLM Evaluation.} We propose an evaluation pipeline for LLM-generated \newtask, with three steps: \textit{(i) Code Generation.} Given a scenario description and prompt, the LLM generates Python code for an animation. The prompt specifies object representations, duration, and MP4 output. \textit{(ii) Video Rendering.} We extract and execute the Python code to render videos depicting qualitative dynamics. \textit{(iii) VLM-based Verification.} We feed the videos and verification questions into a vision–language model.\footnote{We compare open-source VLMs and find Qwen-2.5-72B-VL provides the most accurate judgments.} The model answers each question as \textsc{true} or \textsc{false}, and a response is consistent only if all answers are \textsc{true}. Consistent and inconsistent examples are shown in the supplementary document.

\begin{table}[t]
\centering
\begin{tabular}{p{0.62\linewidth}c}
\toprule  
Model & Agreement (\%) \\
\midrule
Qwen2.5-VL-72B-Instruct & 87.8 \\
Qwen3-VL-235B-A22B-Instruct~\citep{bai2025qwen3vltechnicalreport} & 84.3 \\
InternVL-3.5~\citep{wang2025internvl3_5} & 83.4 \\
GLM-4.6V~\citep{vteam2025glm45vglm41vthinkingversatilemultimodal} & 84.5 \\
\bottomrule
\end{tabular}
\caption{Agreement between VLM and human judges.}
\label{tab:agreement}
\end{table}

\textbf{Validation of Evaluation Reliability.}
To validate the proposed pipeline, we conduct human evaluation on the test set. We use DeepSeek-R1-0528~\citep{deepseekai2025deepseekr1incentivizingreasoningcapability} to generate 8 responses per sample, yielding 2,672 responses, of which 2,136 are successfully rendered. A response passes if the qualitative dynamics in the text are faithfully reproduced in the video. Two independent annotators evaluate these videos, achieving 87\% inter-annotator agreement. Disagreements are resolved through discussion to obtain final human labels.

We compare human labels with VLM-judge outputs. \tabref{tab:agreement} reports agreement rates between VLMs and human annotations. Qwen2.5-VL-72B-Instruct~\citep{qwen2.5-VL} achieves the highest agreement, while all evaluated VLMs exceed 83\%, indicating the reliability of VLM-based evaluation. Error analysis is provided in the supplementary document.

\section{Evaluation of Frontier LLMs}

We evaluate contemporary reasoning models, including open-source models GPT-oss (20B, 120B)~\citep{openai2025gptoss120bgptoss20bmodel}, Qwen3 (32B, 235B-A22B) \citep{qwen3technicalreport}, DeepSeek-R1-0528 \citep{deepseekai2025deepseekr1incentivizingreasoningcapability} and DeepSeek-V3.1 (671B)~\citep{deepseekai2024deepseekv3technicalreport}, as well as proprietary models Gemini-2.5-pro \citep{comanici2025gemini}, GPT-o3, GPT-o4-mini~\citep{openai_o3_o4_mini_2025}, and GPT-5~\citep{openai_gpt5_medium}. 

\textbf{Metrics.} Based on this process, we define the following metrics over all generated responses.
\begin{itemize}
    \item \textbf{Executable rate (E.R.)} The percentage of generated code that is executable. 
    \item \textbf{Render rate (R.R.)} The percentage of generated code that successfully renders videos. 
    \item \textbf{Playable rate (P.R.)} The percentage of generated code that results in a playable video.
    \item \textbf{Accuracy (Acc.)} The percentage of outputs passing all VLM-based verification questions: our primary metric.
\end{itemize}

\textbf{Evaluation.} \tabref{tab:benchmark} summarizes results on \newtask. We report \textbf{Avg@8} and \textbf{Pass@8} to assess average output quality and best-case capability. To probe model capabilities, we employ a carefully designed complex prompt (detailed in the supplementary document), while training uses simpler prompts to encourage better generalization. All models are evaluated under default configurations; unless otherwise specified, results use the thinking-enabled mode, while $^{*}$ marks non-thinking mode. 

\textbf{Low End-to-End Performance Highlights Task Difficulty.} All models perform poorly in end-to-end evaluation, showing limited ability to translate dynamic physical scenes into executable code that produces visually aligned animations. Even DeepSeek-R1-0528 achieves only 21.5\% Avg@8 accuracy, underscoring the difficulty of \newtask and revealing gaps in physics-inspired animation generation.

\begin{table*}[t]

\begin{tabular}{lcrrrcrrrc}
\toprule
\multirow{2}{*}{\textbf{Model}} & \multirow{2}{*}{\textbf{Size}}
& \multicolumn{4}{c}{\textbf{Avg@8}} & \multicolumn{4}{c}{\textbf{Pass@8}} \\
\cmidrule(lr){3-6} \cmidrule(lr){7-10}
& & \textbf{E.R.} & \textbf{R.R.} & \textbf{P.R.} & \textbf{Acc.} & \textbf{E.R.} & \textbf{R.R.} & \textbf{P.R.} & \textbf{Acc.} \\ 
\midrule
Gemini-2.5-pro \citep{comanici2025gemini}              & -          & 99.1       & 77.3              & 38.0      &12.7     & 100.0   & 99.7 & 77.8 & 37.4\\ 
GPT-o3 \citep{openai_o3_o4_mini_2025}                      & -          & 98.9       & 75.2              & 64.1     &  15.9   & 100.0   & 100.0 & 95.5 & 45.2\\ 
GPT-o4-mini \citep{openai_o3_o4_mini_2025}                 & -          & 88.1       & 86.9              & 78.6     &  17.2   & 100.0   & 100.0 & 95.5 & 42.5\\ 
GPT-5-medium \citep{openai_gpt5_medium}                 & -          & 97.2       & 68.9              & 55.7     &  \underline{20.5}   & 100.0   & 100.0 & 93.1 & \textbf{59.9} \\ 
GPT-oss-20b~\citep{openai2025gptoss120bgptoss20bmodel}$^{*}$                 & 21.5B      & 99.1       & 92.0              & 61.6     &  10.5    & 100.0   & 100.0 & 85.6  & 32.0\\
GPT-oss-120b~\citep{openai2025gptoss120bgptoss20bmodel} $^{*}$            & 120B &99.8   &  93.7  &  76.0  &  14.0  &  100.0   &   99.7    &    91.9    & 37.4   \\ 
Qwen3-32B \citep{qwen3technicalreport}                   & 32B        & 99.8       & 96.6              & 51.0     &  11.1   & 100.0   & 100.0 & 82.6 & 30.5\\
Qwen3-235B-A22B \citep{qwen3technicalreport}             & 235B       & 99.7       & 95.9              & 68.5     &  15.1   & 100.0   & 100.0 & 91.3 & 41.0\\
DeepSeek-R1-0528 \citep{deepseekai2025deepseekr1incentivizingreasoningcapability}            & 671B     & 99.7       & 97.3             & 79.9     &  \textbf{21.5}   & 100.0   & 100.0 & 93.4 & \underline{52.7} \\
DeepSeek-V3.1\citep{deepseekai2024deepseekv3technicalreport}$^{*}$                & 671B       & 99.9       & 95.2              & 64.0     &  14.5    & 100.0   & 99.4 & 90.4 & 40.7\\ 

\bottomrule
\end{tabular}
\caption{Comparison of frontier LLMs on \newtask. Metrics include Executable Rate (E.R.), Render Rate (R.R.), Playable Rate (P.R.), and VLM-evaluated Accuracy (Acc.). Avg@8 reports average performance over 8 runs, while Pass@8 counts a scenario as correct if any of the 8 responses pass all verification questions. Rows marked with $^{*}$ indicate non-thinking mode.}
\label{tab:benchmark}
\end{table*}

\begin{table}[t]
\centering
\begin{tabular}{lrr}
\toprule  
Category & Avg@8 & Pass@8 \\
\midrule
Mechanics        & 27.7 & 58.8 \\
Electromagnetism & 20.8 & 60.0 \\
Optics           & 8.4  & 27.0 \\
Fluid Mechanics  & 20.5 & 50.0 \\
Thermodynamics   & 26.0 & 80.0 \\
\bottomrule
\end{tabular}
\caption{DeepSeek-R1-0528 breakdown across domains.}
\label{tab:categories}
\end{table}

\textbf{High Code Generation Capability but Weak Scene Alignment.} Most models achieve high executable rate (E.R.) and render rate (R.R.), showing that they can generate video-producing code. However, lower playable rate (P.R.) and accuracy (Acc.) indicate persistent difficulties in translating scene descriptions into visually aligned animations with plausible qualitative dynamics.

\textbf{Pass@8 Reveals Hidden Potential.} While Avg@8 is low across all models, Pass@8 shows higher success rates, with the best models approaching ~60\%. This gap indicates that models can sometimes generate code whose animations pass visual verification, but lack consistency across attempts.

\tabref{tab:categories} reports domain-level performance. DeepSeek-R1-0528 varies across domains, performing best on mechanics (27.7\% Avg@8) and thermodynamics (80.0\% Pass@8). Optics is the most challenging domain, with only 8.4\% Avg@8 and 27.0\% Pass@8, reflecting scarce optics-related supervision during training. Nevertheless, low Avg@8 scores across all domains highlight the difficulty of generating aligned animations that capture qualitative physical dynamics.

\section{Model Training for \newtask}

In this section, we present experiments and results for the proposed training approach, which applies SFT followed by RL with a vision-based reward signal using the curated training data. We introduce the experimental setup, reward design, main results, and reward-hacking analysis.

\subsection{Experimental Setup} 
For supervised fine-tuning, we use \newtask data to fine-tune DeepSeek-R1-Distill-Qwen-7B and Qwen3-32B. We use AdamW with a maximum learning rate of $4\times 10^{-5}$, linearly decayed to 0. Models are trained for 2 epochs with a batch size of 256, and samples are truncated to a maximum of $16,384$ tokens. Training takes about 32 GPU hours in total.

For RL, we adopt GRPO~\citep{shao2024deepseekmathpushinglimitsmathematical} to aggregate reward implemented on top of AgentFly~\citep{wang2025agentfly} (see the supplementary document for additional training details). We use the Adam optimizer with a learning rate of $5\times10^{-7}$ and train for $150$ steps with a global training batch size of $32$ and $8$ rollouts per sample, totaling $576$ GPU-hours. All rollouts are truncated to a maximum of $16,384$ tokens.

\subsection{Code-Video-Judge Reward Design}
Each sample scenario $s$ in \newtask is paired with a set of verification questions $\mathcal{Q}=\{q_i\}_{i=1}^{M}$, designed to test whether the rendered animation is visually aligned with the scenario description and captures the expected qualitative dynamics. For each $q_i$, the VLM yields a binary label $y_i\in\{\textsc{true},\textsc{false}\}$. If compilation or execution fails, we set $\tilde{y}_i=\textsc{false}$ for all $i$; otherwise, $\tilde{y}_i=y_i$. The binary reward is $r_{\mathrm{binary}}=\mathbf{1}[\tilde{y}_i=\textsc{true},\ \forall i\in\{1,\ldots,M\}]$, meaning that all verification questions must be answered \textsc{true}.

\subsection{Main Results}
\begin{table}[t]
\centering
\small
\setlength{\tabcolsep}{3.5pt}
\begin{tabular}{lrrrrrrrr}
\toprule
\multirow{2}{*}{Model} 
& \multicolumn{4}{c}{Avg@8} 
& \multicolumn{4}{c}{Pass@8} \\
\cmidrule(lr){2-5} \cmidrule(lr){6-9}
& E.R. & R.R. & P.R. & Acc. & E.R. & R.R. & P.R. & Acc. \\ 
\midrule
Base-7B & 62.1 & 2.3  & 0.2  & 0.0  & 99.1  & 10.5  & 1.8  & 0.0  \\
SFT-7B  & 93.8 & 87.1 & 30.0 & 4.4  & 100.0 & 100.0 & 72.8 & 18.3 \\
RL-7B   & 99.1 & 98.1 & 81.5 & 11.1 & 100.0 & 100.0 & 95.2 & 34.4 \\
\midrule
Base-32B & 99.8 & 96.6 & 51.0 & 11.1 & 100.0 & 100.0 & 82.6 & 30.5 \\
SFT-32B  & 97.7 & 95.3 & 74.3 & 17.4 & 100.0 & 100.0 & 92.0 & 43.1 \\
RL-32B   & 99.9 & 97.7 & 90.6 & 36.9 & 100.0 & 100.0 & 98.5 & 72.2 \\
\bottomrule
\end{tabular}
\caption{Performance of 7B and 32B models across base, SFT, and RL stages.}
\label{tab:training}
\end{table}
We report results for \emph{DeepSeek-R1-Distill-Qwen-7B} and \emph{Qwen3-32B} trained with our SFT and RL pipelines. Unless otherwise stated, we use the VLM-as-judge binary reward $r_{\mathrm{binary}}$. Overall results are summarized in \tabref{tab:training}. The base model struggles with physics-inspired animation generation, while SFT improves all metrics substantially.

Compared to SFT-only training, RL boosts render-to-play conversion (P.R./R.R.), improving runtime reliability and pipeline assembly. In terms of Pass@8, under RL, performance approaches saturation with E.R.\ = 100.0\%, R.R.\ = 100.0\%, and P.R.\ = 95.2\%, while Acc.\ rises from 18.3\% to 34.4\% for a gain of about 1.8$\times$.

\begin{figure}[t]
  \centering
  \includegraphics[width=.9\linewidth]{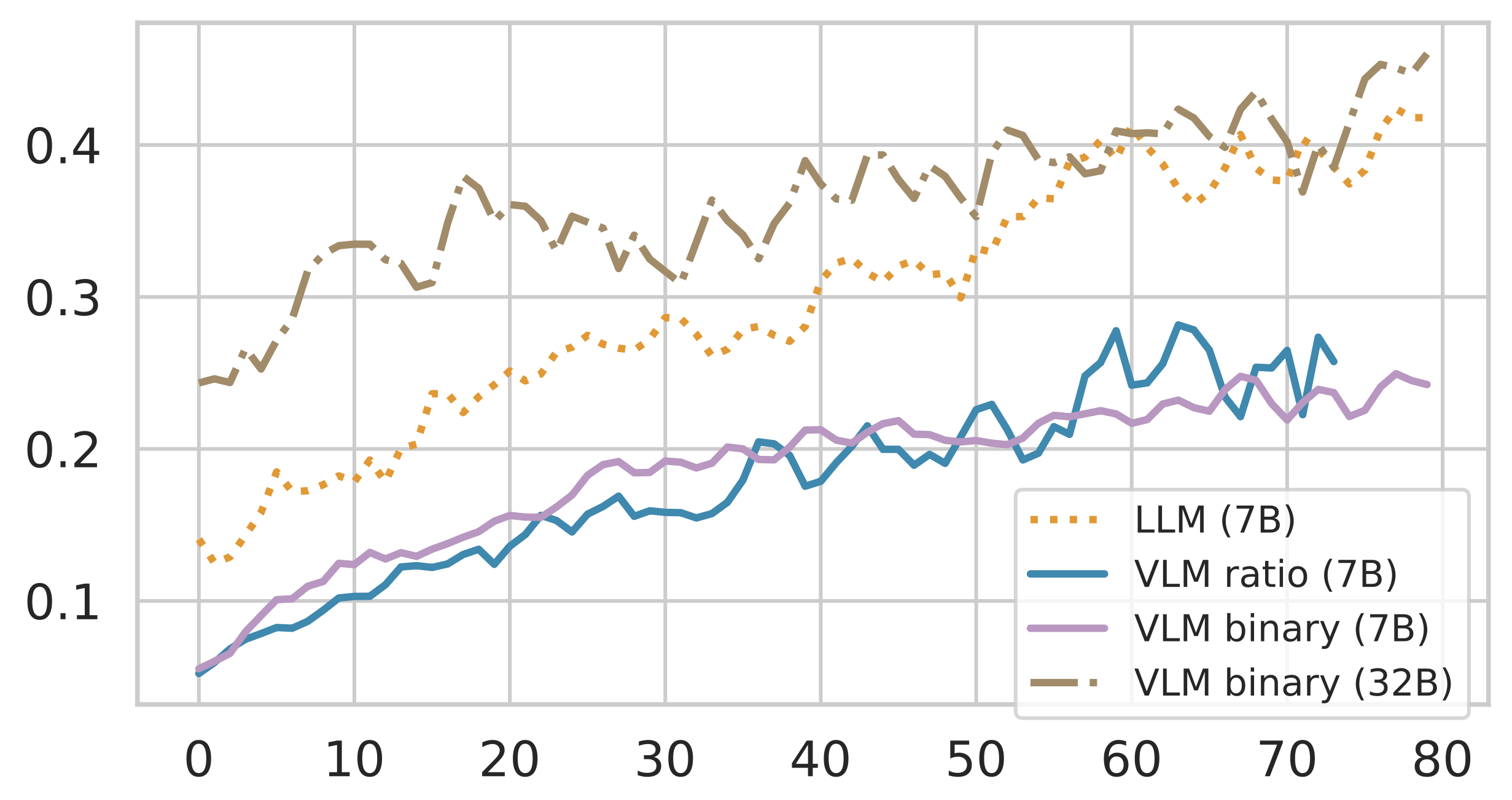}
  \caption{Reward curves during RL training, showing steady improvement across rewards.}
  \label{fig:training_curve}
\end{figure}

With training, the 7B model attains \textbf{34.4\%} Pass@8 accuracy, comparable to several larger models. The 32B model further achieves the best performance (Pass@8 \textbf{72.2\%}), surpassing strong untrained models such as DeepSeek-R1. These gains show that targeted data curation and training meaningfully improve physics-inspired animation generation, supporting our dataset design, training procedure, and vision-based reward signal. To isolate whether RL improves visual alignment beyond execution, rendering, and playability, we conduct a controlled subset analysis in the supplementary document; results show that RL improves accuracy even when upstream pipeline stages are controlled.

\subsection{Ablation Studies} 
\textbf{Pipeline-Stage Ablation.}
We compare base, SFT, and RL models in \tabref{tab:training} to isolate the effects of curated \newtask data and the Code-Video-Judge reward. SFT substantially improves all metrics over the base model, indicating that curated \newtask provides useful supervision. RL further improves end-to-end performance, especially render-to-play conversion and final accuracy, showing that the Code-Video-Judge reward contributes beyond supervised learning.

\textbf{Reward Ablation.}
 We compare the proposed \emph{Code-Video-Judge} binary reward, labeled \emph{VLM binary} in the figure, with two reward variants. The first uses the same VLM-based video judging mechanism but replaces the binary all-questions-pass criterion with a ratio reward, defined as the fraction of correctly verified questions between 0 and 1:
$
r_{ratio} \;=\; \frac{1}{M}\sum_{i=1}^{M}\mathbf{1}\!\left[\tilde{y}_i=\textsc{true}\right]
$.
The second variant uses \emph{LLM-as-judge} (text-only) reward, where the generated code is judged directly without videos or applying visual verification questions. Details are provided in the supplementary document. Training curves are shown in \figref{fig:training_curve}.

As shown in \figref{fig:training_curve}, all judges provide steadily improving signals during RL. Although the \emph{VLM-as-judge} variants do not achieve higher scores than \emph{LLM-as-judge}, they yield better end-to-end performance, indicating more faithful and consistent rewards. We attribute this gap to two factors. \textbf{(i) Modality fidelity.} VLMs evaluate rendered videos and observe motion and interactions, whereas text-only LLMs infer runtime behavior from static code, leading to optimistic false positives for physically implausible or visually invalid outcomes. \textbf{(ii) Execution observability.} We assign zero reward to compilation or execution failures, coupling VLM rewards to end-to-end reliability. In contrast, text-only judges do not execute code and may overestimate superficially valid but non-rendering code.

Thus, visual feedback closes the code-to-video-to-verification loop and aligns rewards with the end-to-end objective. Response length trends appear in the supplementary document.

\subsection{Discussion and Reward Hacking}
Reward hacking refers to the policy exploiting biases, artifacts, or loopholes in the reward function to achieve high rewards without genuinely improving the true objective. Since our rewards are based on VLM judgments of videos, we discuss mitigation strategies and conduct manual checks to assess whether such behavior occurs during training.

\textbf{Preventing Reward Hacking.}
We use three strategies to mitigate reward hacking: (1) \textit{Multi-question aggregated rewards.} We evaluate 4--12 verification questions per scenario and aggregate their scores, making single-question reward exploits harder. (2) \textit{Ensemble VLM voting during training.} Three VLMs evaluate each video, with majority voting determining the final reward. (3) \textit{Train-test VLM separation.} We train with Qwen3-VL, InternVL3.5, and GLM-4.6V, but evaluate with Qwen2.5-VL-72B-Instruct. Together with human evaluation, this separation provides evidence against systematic reward hacking.

\textbf{Empirical Validation Through Human Evaluation.}
To further verify that our RL-trained models achieve improvements rather than exploiting VLM biases, we conduct human evaluation on model outputs across training stages. Due to the annotation cost, approximately 30 hours per model, we perform stage-wise human evaluation for the 7B model across the base, SFT, and RL stages. Human-VLM agreement remains high across all stages: 100.0\% for the base model, 88.2\% for the SFT model, and 88.7\% for the RL model. The 100.0\% agreement at the base stage is expected, since the base model produces very few playable videos and its failures are typically clear-cut and unambiguous. The consistently high agreement at the SFT and RL stages confirms that the observed improvements reflect genuine gains in animation quality rather than artifacts of the VLM judge.

For the 32B model, we observe 86.9\% human-VLM agreement at the RL stage. This evaluation includes failure cases, as incorrect or visually inconsistent animations are assessed by VLMs. The high agreement between human and VLM judgments suggests that such cases are reliably identified, with no clear evidence of reward hacking under our training setup. We further conduct a direct side-by-side comparison on the same scenario in the supplementary document: the base model fails to generate a playable video, the SFT model generates a video but misses the key motion, and the RL model produces a coherent animation that follows the scenario description. This progression demonstrates that improvements are substantive and reflect genuine gains in text-to-code-to-video consistency rather than shortcutting the VLM judge.}

\section{Conclusion}

We introduced \newtask, a new task, dataset, and benchmark for evaluating LLMs' capabilities on generating executable code that produces physics-inspired animations aligned with textual physical scenarios. We built a fully automatic pipeline that \textit{(i)} generates topic-conditioned dynamic scenarios and visual verification questions, \textit{(ii)} produces reasoning traces and executable code, and \textit{(iii)} validates rendered videos via a VLM. The dataset contains 7,659 scenarios across five physics domains, with a 334-example human-verified test set; VLM judgments show 87.8\% agreement with human labels, supporting their use as reliable evaluators. Results for contemporary LLMs reveal that the task is challenging, highlighting gaps in current models' ability to generate visually aligned animations that capture qualitative physical dynamics. Building on \newtask, we show that training LLMs via VLM-based rewards derived from code-generated videos leads to improved end-to-end text-to-code-to-video consistency.

\bibliography{aaai2027}

\clearpage
\appendix
\onecolumn
\section{Additional Related Work and Task Positioning}

\paragraph{Physics Reasoning, Simulation, and Visual Verification.}
Existing physics reasoning benchmarks primarily evaluate discriminative or numerically grounded capabilities. For example, UGPhysics~\citep{xu2025ugphysicscomprehensivebenchmarkundergraduate} and PhysReason~\citep{zhang2025physreason} focus on solving physics problems or verifying numerical correctness, while SeePhys~\citep{xiang2025seephysdoesseeinghelp} studies diagram-centric, vision-essential physics QA. These benchmarks cover an important but orthogonal task dimension: whether models can reason over given problems, diagrams, or numerical answers. In contrast, \newtask evaluates a generative capability: producing executable code whose rendered animation is visually aligned with a natural-language physical scenario description and captures the expected qualitative dynamics.

This distinction matters because \newtask requires a combination of physical-scenario understanding, animation-oriented code generation, and text-to-code-to-video consistency. Existing numerically grounded benchmarks typically assess problem solving, answer verification, or reasoning against ground-truth trajectories and calibrated parameters. By contrast, our setting asks whether a model can translate a qualitative physical scenario into executable code whose rendered animation visually depicts the expected motion process.

By using fine-grained visual verification questions over rendered animations, our pipeline avoids reliance on ground-truth trajectories or simulator-specific reference outputs. This makes it scalable to diverse scenarios across 52 physical concepts and fills a complementary gap in physics-inspired animation generation.

More broadly, \newtask reflects a general code-to-visual-verification paradigm: an LLM generates executable code, the code is rendered, and a VLM evaluates functional and perceptual consistency. Physical scenarios serve as a testbed because their qualitative dynamics are visually observable, but the same framework can extend to educational visualization, embodied-agent environments, procedural animation, and CAD/3D design verification. Recent work supports this paradigm: VLM-based RL feedback (RL-VLM-F \citep{wang2024rl}), general-purpose VLM evaluators (Prometheus-Vision \citep{lee2024prometheus}), and code-to-visual verification (CADCodeVerify \citep{alrashedy2025generating}) demonstrate that multimodal VLM judgments can effectively supervise complex, dynamic, and structured outputs.

\section{Dataset Details}
\subsection{Domain and Concept Distribution}
We follow the train/test and SFT/RL splits described in the main paper, and provide additional statistics on the domain and concept distributions below. We use a corpus of \textbf{7{,}659} samples covering five core physics domains with \textbf{52} fine-grained concepts (see \tabref{tab:domains}). Mechanics contributes the largest share (\(3{,}533;\,46.1\%\)), followed by Electromagnetism (\(1{,}997;\,26.0\%\)), Optics (\(840;\,11.0\%\)), Fluid Mechanics (\(724;\,9.5\%\)), and Thermodynamics (\(565;\,7.4\%\)). Across concepts, the mean count is \(147.3\) samples (median \(149.5\); s.d. \(82.6\)), indicating a moderately imbalanced, long-tailed distribution.

At the concept level, the distribution exhibits a long tail within each domain: a few concepts have many examples, while many have far fewer. \emph{Fluid Dynamics} is the largest category (\(519;\,6.8\%\)), with other frequent topics including \emph{Angular Momentum} (\(259;\,3.4\%\)), \emph{Gravity} (\(252;\,3.3\%\)), and \emph{Electric Potential} (\(240;\,3.1\%\)). On the sparse end, \emph{Inductance} (\(8;\,0.1\%\)), \emph{Polarization} (\(10;\,0.1\%\)), \emph{Torque} (\(13;\,0.2\%\)), \emph{Coulomb's Law} (\(16;\,0.2\%\)), and \emph{Simple Pendulum} (\(16;\,0.2\%\)) have very few examples. The ten most common concepts together account for \(33.7\%\) of the data, whereas the ten rarest account for \(5.9\%\).

\begin{table*}[t]
\centering   
\small
\setlength{\tabcolsep}{4pt}     
\begin{tabular*}{\textwidth}{@{\extracolsep{\fill}} l l r l r @{}}
\toprule
\textbf{Domain} & \textbf{Concept} & \textbf{Number} & \textbf{Concept} & \textbf{Number}\\
\midrule
\multirow{8}{*}{Electromagnetism}
 & Electric Potential & 240 & Magnetic Field & 200\\
 & Electric Field & 178 & Capacitance & 170\\
 & Gauss's Law & 169 & Ohm's Law & 155\\
 & Resistance & 149 & Mutual Inductance & 128\\
 & Faraday's Law & 127 & Ampère's Law & 124\\
 & Magnetic Flux & 94 & Electric Current & 90\\
 & Lenz's Law & 79 & Self-Inductance & 70\\
 & Coulomb's Law & 16 & Inductance & 8\\
 & \multicolumn{3}{r}{\textbf{Total}} & \textbf{1997}\\
\midrule
\multirow{1}{*}{Fluid Mechanics}
 & Fluid Dynamics & 519 & Fluid Statics & 205\\
 & \multicolumn{3}{r}{\textbf{Total}} & \textbf{724}\\
\midrule
\multirow{12}{*}{Mechanics}
 & Angular Momentum & 259 & Gravity & 252\\
 & Moment of Inertia & 234 & Conservation of Momentum & 213\\
 & Potential Energy & 208 & Linear Momentum & 194\\
 & Elastic Collision & 181 & Linear Motion & 180\\
 & Kinetic Energy & 178 & Friction & 172\\
 & Damped Oscillation & 159 & Rotational Motion & 155\\
 & Spring Oscillator & 150 & Driven Oscillation & 126\\
 & Relative Motion & 125 & Conservation of Mechanical Energy & 115\\
 & Circular Motion & 110 & Curvilinear Motion & 107\\
 & Impulse & 104 & Inelastic Collision & 102\\
 & Elastic Force & 92 & Resonance & 88\\
 & Simple Pendulum & 16 & Torque & 13\\
 & \multicolumn{3}{r}{\textbf{Total}} & \textbf{3533}\\
\midrule
\multirow{3}{*}{Optics}
 & Refraction of Light & 226 & Lens & 210\\
 & Reflection of Light & 153 & Interference of Light & 143\\
 & Diffraction of Light & 98 & Polarization & 10\\
 & \multicolumn{3}{r}{\textbf{Total}} & \textbf{840}\\
\midrule
\multirow{2}{*}{Thermodynamics}
 & Temperature & 201 & Ideal Gas Law & 184\\
 & Pressure & 107 & Heat & 73\\
 & \multicolumn{3}{r}{\textbf{Total}} & \textbf{565}\\
\midrule
\multicolumn{4}{r}{\textbf{Grand total}} & \textbf{7659}\\
\bottomrule
\end{tabular*}
\caption{Distribution of samples across domains and concepts.}
\label{tab:domains}
\end{table*}
We intentionally include both well-represented concepts (e.g., Angular Momentum, Electric Potential, Refraction of Light) and low-frequency ones (e.g., Inductance, Polarization, Torque) to ensure topic diversity. The resulting long-tail distribution reflects differences in how readily concepts can be represented as visual dynamic scenarios, while still allowing the benchmark to evaluate LLMs across both common and visually subtle physical concepts.

\subsection{Analysis of Dataset Properties}
In this section, we further analyze properties of the \textbf{\newtask} dataset beyond the domain and concept table, including the source of domain imbalance, scenario lengths, and verification question counts.

\textbf{Domain Imbalance.} The distribution across the five high-level domains is intentionally non-uniform rather than uniformly sampled. This is because the domains differ in the number of visually representable concepts and in how naturally their concepts can be expressed as dynamic code-generated animations. Mechanics includes 24 core concepts and electromagnetism includes 16, together comprising the majority of the dataset. The remaining three domains account for only 12 core concepts in total. We therefore include more examples from mechanics and electromagnetism, whose concepts are more common in everyday physical interactions and generally easier to visualize as qualitative motion patterns.

\begin{figure*}[t]
  \centering

  \includegraphics[width=0.58\textwidth]{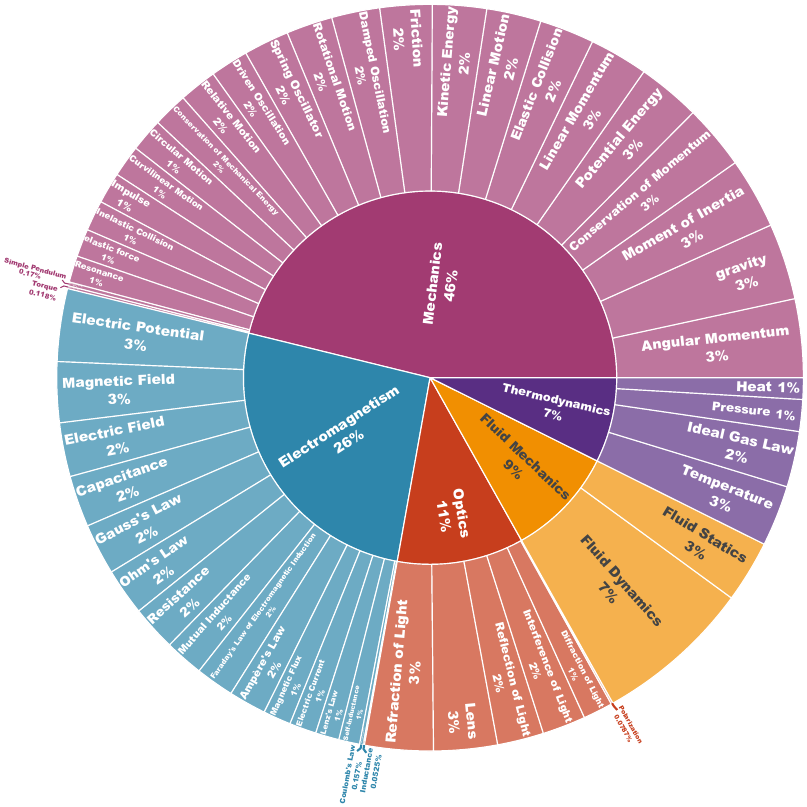}
  \par\vspace{2pt}
  \textbf{(a)} Topic distribution of generated scenarios across physics domains.

  \vspace{0.8em}

  \begin{minipage}[t]{0.4\textwidth}
    \centering
    \includegraphics[width=\linewidth]{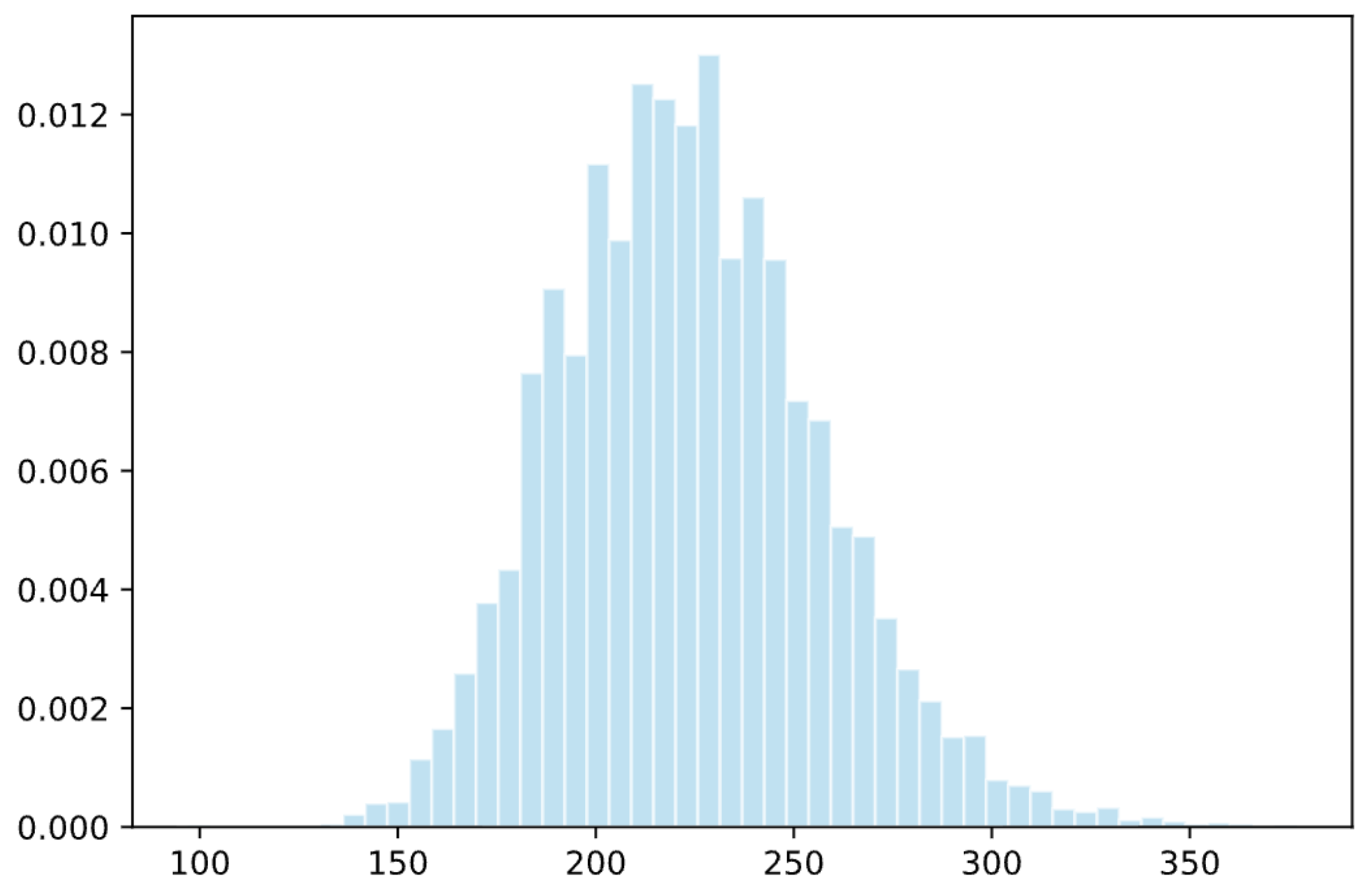}
    \par\vspace{2pt}
    \textbf{(b)} Token count distribution per scenario.
  \end{minipage}
  \hfill
  \begin{minipage}[t]{0.4\textwidth}
    \centering
    \includegraphics[width=\linewidth]{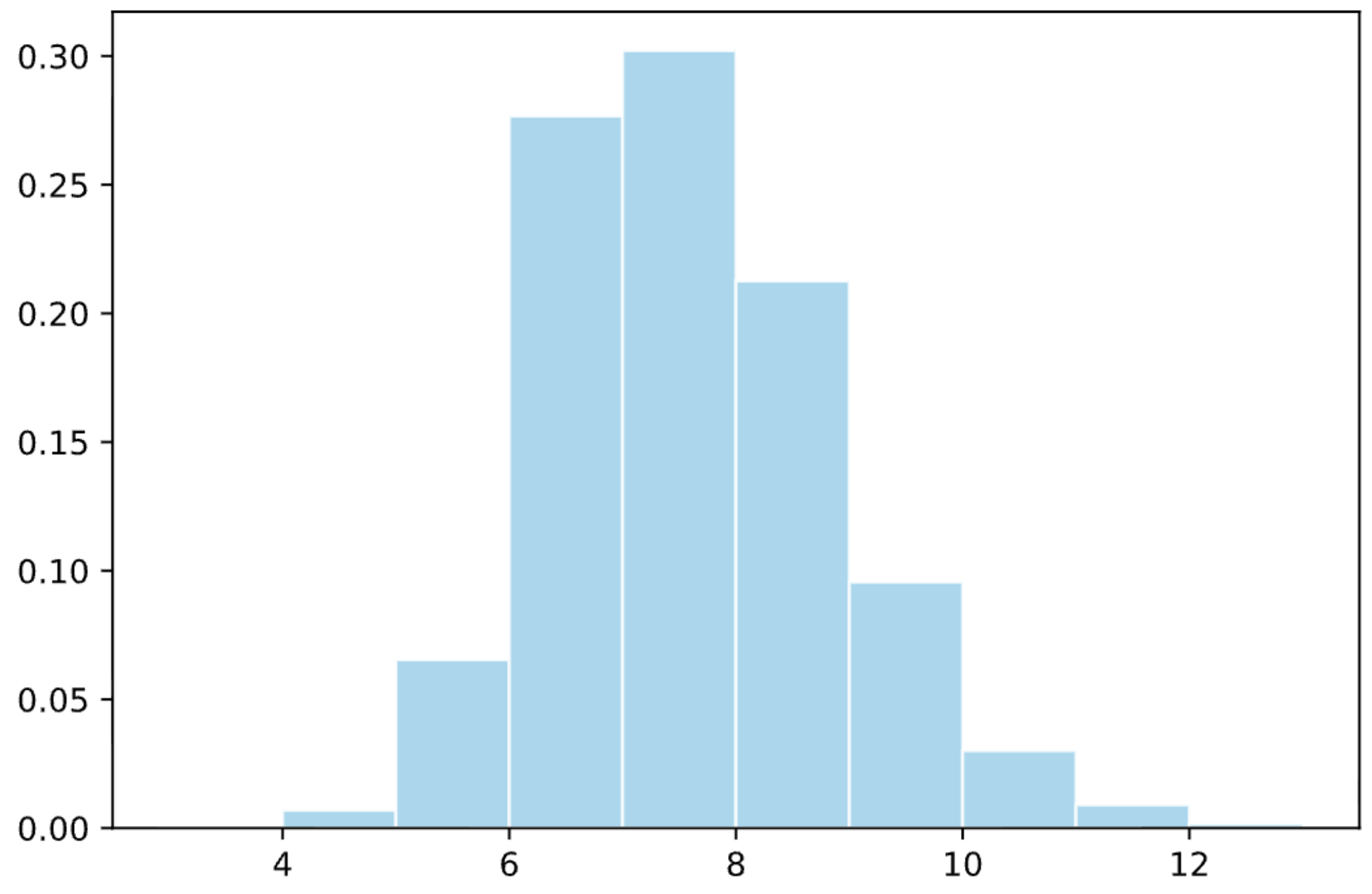}
    \par\vspace{2pt}
    \textbf{(c)} VLM verification question count distribution.
  \end{minipage}
  \caption{Distributions of topics, scenario description token counts, and verification question counts in the \newtask dataset. Panel (a) shows the topic distribution across physics domains and concepts, where the inner ring denotes domains and the outer ring denotes concepts. Panel (b) shows token counts per scenario measured using the Qwen3 tokenizer. Panel (c) shows the number of verification questions per scenario.}
  \label{fig:more_stats}
\end{figure*}

\textbf{Scenario Length and Question Counts.} Beyond domains, we examine both the length of generated scenarios and the number of associated verification questions. Using the Qwen3 tokenizer (\figref{fig:more_stats}(b)), scenario lengths range from 100 to 350 tokens, with a near-normal distribution centered around 225, suggesting moderate complexity while avoiding extremes of brevity or verbosity. For verification questions (\figref{fig:more_stats}(c)), each scenario yields between 4 and 12 questions, most often 6 to 10, which provides coverage of the expected qualitative dynamics without excessive redundancy while preserving scalability across varying levels of complexity.

\section{Evaluation Reliability}
\subsection{Human Verification Protocol}
Human verification was conducted by two PhD students with expertise in computer vision. The assessment of physical plausibility focused on qualitative and macroscopically observable motion patterns relevant to our task, rather than numerical accuracy, equation-level verification, or expert-level physical analysis. This annotation scope does not require specialized training in physics and can be reliably assessed from the scenario descriptions and rendered videos. The verification served two purposes: validating the quality of the test scenarios and verification questions, and assessing whether automatic judge predictions correctly reflected the rendered videos.

Specifically, annotators assessed: \textit{(i)} whether each scenario description is physically plausible and unambiguous; \textit{(ii)} whether the associated verification questions adequately cover the expected qualitative dynamics without excessive redundancy; and \textit{(iii)} whether judge predictions correctly reflect the content of the rendered video after aggregating the verification questions. This process was rigorous rather than a quick plausibility check. A complete verification of the full test set took approximately 30 hours per annotator. These annotations were used to validate both the quality of the benchmark and the reliability of VLM-based evaluation.

\subsection{VLM-as-Judge versus LLM-as-Judge}

We compare \emph{VLM-as-judge} and \emph{LLM-as-judge} to highlight the importance of visual verification. Both judges use the same DeepSeek-R1-0528 generated responses and the same verification questions, but differ in the evidence available to the judge. In the VLM-as-judge setting, the model receives the rendered video, whereas in the LLM-as-judge setting, it receives only textual inputs and is used as a text-only code judge.

For \emph{VLM-as-judge}, we execute the generated code, render the corresponding video, and pass the video together with the verification questions to Qwen2.5-VL-72B-Instruct~\citep{qwen2.5-VL}. The VLM answers each question based on the rendered video. For the \emph{LLM-as-judge} setting, we send the generated code directly to Qwen2.5-VL-72B-Instruct~\citep{qwen2.5-VL}, together with the visual verification questions. To approximate the execution and video-generation checks available in the VLM-based pipeline, we additionally include two text-only questions asking whether the code appears executable and whether it appears to generate and save a video. The LLM judges all questions as \textsc{True} or \textsc{False} based only on the code. This variant does not execute the code, render a video, or inspect the visual output. In both settings, an item is considered correct only if all questions used by the corresponding judge are answered as \textsc{True}: the original visual verification questions for \emph{VLM-as-judge}, and the visual verification questions plus the two text-only execution and video-generation checks for \emph{LLM-as-judge}.

\subsection{Agreement with Human Annotation}
We evaluate both judges on the same DeepSeek-R1-0528 responses and compare their predictions against human annotations. Each test scenario is evaluated with eight generated responses. Each response is rendered into a video, and the judge answers all verification questions for that video. The response-level prediction is obtained by aggregating these question-level judgments using the all-true rule.

\tabref{tab:judge_results} reports the results on the test set. \textit{True/False} means the judge predicts \textsc{true} while the human annotator labels the example as \textsc{false}, and \textit{False/True} means the judge predicts \textsc{false} while the human annotator labels the example as \textsc{true}. \textit{Agreement} reflects the consistency between judge predictions and human annotations. As shown, \emph{VLM-as-judge} achieves markedly higher agreement with human annotations (87.8\%) than \emph{LLM-as-judge} (69.1\%), demonstrating the benefit of incorporating code execution and video rendering into the evaluation process.

\begin{table*}[t]
\centering
\begin{tabular}{lccccc}
\toprule
 & True/True & False/False & True/False & False/True& Agreement \\ \midrule
VLM-as-judge & 613 & 1263 & 108 & 152 & 87.8\% \\
LLM-as-judge & 266 & 1209 & 162 & 499 & 69.1\% \\ \bottomrule
\end{tabular}
\caption{Evaluation results under different judges. 
The table compares judge predictions with human annotations. 
\textit{True/False} means the judge predicts \textsc{true} while the human annotator labels the example as \textsc{false}; 
\textit{False/True} means the judge predicts \textsc{false} while the human annotator labels the example as \textsc{true}. 
\textit{Agreement} indicates consistency between judge predictions and human annotations.}
\label{tab:judge_results}
\end{table*}

To further validate the reliability of VLM-based evaluation during training, we report human-evaluated pass rates alongside VLM pass rates for the 7B model at all three training stages in \tabref{tab:human_vlm_rl}. The ranking is preserved under human evaluation, with Base $<$ SFT $<$ RL, confirming that VLM scores reflect genuine improvements rather than artifacts of the evaluation pipeline. Notably, human pass rates are consistently slightly higher than VLM pass rates, suggesting that the VLM scores are not systematically inflated. This provides additional evidence against systematic reward hacking or exploitation of VLM-specific visual cues.

\begin{table*}[t]
\centering
\begin{tabular}{lccc}
\toprule
Model stage & VLM pass rate & Human pass rate & Human--VLM agreement \\
\midrule
Base & 0.0\% & 0.0\% & 100.0\% \\
SFT  & 4.4\% & 6.1\% & 88.2\% \\
RL   & 11.1\% & 15.8\% & 88.7\% \\
\bottomrule
\end{tabular}
\caption{Human validation of VLM-based evaluation for the 7B model across training stages. Human pass rates preserve the same ranking as VLM pass rates, while slightly higher human pass rates suggest that the VLM judge is conservative rather than over-permissive.}
\label{tab:human_vlm_rl}
\end{table*}

\subsection{VLM Evaluation Error Analysis}
\begin{figure*}[t]
  \centering
  \includegraphics[width=\textwidth]{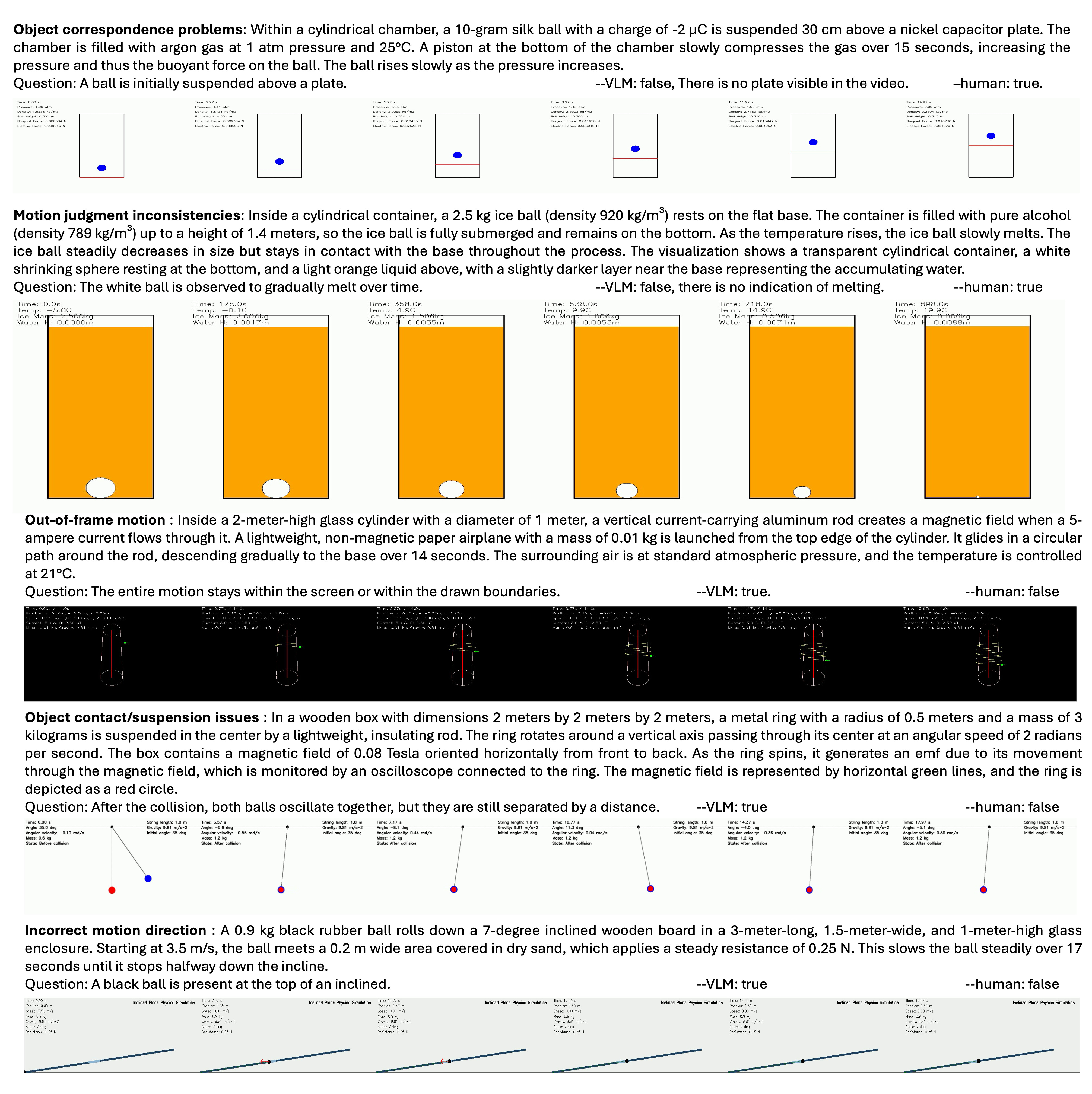}
  \caption{Representative disagreement between Qwen2.5-VL-72B-Instruct and human. }
  \label{fig:data_humanvalid}
\end{figure*}

To better understand the limitations of VLM-based evaluation, we analyze the 252 disagreement cases between Qwen2.5-VL-72B-Instruct and human annotations. We identify five main error categories: \textit{Object correspondence problems (36\%):} the VLM incorrectly matches objects in the video to those described in the text; \textit{Qualitative dynamics judgment inconsistencies (35\%):} the VLM's assessment of motion characteristics conflicts with human observations; \textit{Out-of-frame motion (19\%):} critical motion occurs outside the visible frame, while the VLM incorrectly judges it as visible within the frame; \textit{Object contact/suspension issues (6\%):} the VLM fails to correctly identify whether objects are in contact or improperly suspended; and \textit{Incorrect motion direction (4\%):} the VLM misjudges the direction of object movement. Representative examples are shown in \figref{fig:data_humanvalid}.

\section{Training and Reward Details}
\subsection{GRPO Optimization}
We adopt Group Relative Policy Optimization (GRPO)~\citep{shao2024deepseekmathpushinglimitsmathematical}. For each prompt $x$, the old policy $\pi_{\theta_{\mathrm{old}}}$ samples a \emph{group} of $K$ candidates $c_{1:K}\sim \pi_{\theta_{\mathrm{old}}}(\cdot\,|\,x)$ (e.g., $K=8$). Executing each $c_k$ produces a reward $r_k=r(c_k\,|\,x)$ from VLM-as-judge binary reward. We compute a group-relative, length-invariant advantage
\begin{equation}
\label{eq:group-adv}
\begin{aligned}
\bar{A}_k &=
\frac{r_k-\mu_x}{\sigma_x+\epsilon}, \\
\mu_x &= \frac{1}{K}\sum_{j=1}^{K} r_j, \\
\sigma_x &=
\sqrt{\frac{1}{K}\sum_{j=1}^{K}(r_j-\mu_x)^2}.
\end{aligned}
\end{equation}
and assign $\bar{A}_k$ uniformly to all tokens of $c_k$ to avoid length bias.

We then optimize a clipped likelihood-ratio objective \emph{without} an explicit KL penalty, using asymmetric clipping to stabilize updates while preserving headroom for improvement:
\begin{equation}
\label{eq:grpo-objective}
\begin{aligned}
\mathcal{L}_{\mathrm{clip}}(\theta)
&= \frac{1}{K}\sum_{k=1}^{K}
   \frac{1}{|c_k|}\sum_{t=1}^{|c_k|}
   \ell_{k,t}(\theta), \\
\ell_{k,t}(\theta)
&= \min \Bigl(
\rho_{k,t}(\theta)\bar{A}_k,\,
\hat{\rho}_{k,t}(\theta)\bar{A}_k
\Bigr), \\
\hat{\rho}_{k,t}(\theta)
&= \operatorname{clip}\!\left(
\rho_{k,t}(\theta),
1-\varepsilon_{\mathrm{low}},
1+\varepsilon_{\mathrm{high}}
\right).
\end{aligned}
\end{equation}
where
\begin{equation}
\label{eq:rho-clip}
\begin{aligned}
\rho_{k,t}(\theta)
&=
\frac{
\pi_{\theta}\!\left(c_{k,t}\mid x, c_{k,<t}\right)
}{
\pi_{\theta_{\mathrm{old}}}\!\left(c_{k,t}\mid x, c_{k,<t}\right)
}, \\
\operatorname{clip}(u,a,b)
&=
\min\!\bigl(\max(u,a),\,b\bigr).
\end{aligned}
\end{equation}
We maximize $\mathcal{L}_{\mathrm{clip}}$ with Adam; $\varepsilon_{\mathrm{low}}{<}\varepsilon_{\mathrm{high}}$ (asymmetric clipping) curbs overly aggressive down-weighting while allowing moderate up-weighting of promising tokens.

\subsection{Reward Design for RL Training}
We provide additional implementation details for the reward functions used in RL training. The main paper defines the \emph{VLM-as-judge} ratio reward used in the reward ablation; here, we focus on the default \emph{VLM-as-judge} binary reward and the text-only \emph{LLM-as-judge} baseline. For VLM-based rewards, if compilation or execution fails, we set all resolved labels to $0$.

\paragraph{(1) VLM-as-judge: binary (pass/fail) reward.}
Using the same $\mathcal{Q}_x$ and resolution rule, we define a stricter binary criterion in which a sample is marked as \emph{pass} only if all resolved answers are positive:
\begin{equation}
r_{\mathrm{binary}}^{\mathrm{VLM}}(c\,|\,x)
\;=\;
\mathbf{1}\!\left[\forall\,i\in\{1,\ldots,M\},\ \tilde y_i=1\right].
\label{eq:reward-pass-vlm}
\end{equation}
This reward provides a sparse but high-precision supervision signal and is used as our default VLM-based reward unless otherwise specified.

\paragraph{(2) LLM-as-judge (text-only): pass/fail reward.}
Here we do \emph{not} execute $c$ nor render a video. Instead, we prompt a text-only LLM with \textit{(i)} the scenario description $x$, \textit{(ii)} the generated code $c$, and \textit{(iii)} an extended question set
\[
\mathcal{Q}'_x \;=\; \mathcal{Q}_x \;\cup\; \{q_{\text{exec}}, q_{\text{video}}\},
\]
where the two additional text-only checks are:
\begin{itemize}
\item $q_{\text{exec}}$: \emph{``Does the code execute without errors?''}
\item $q_{\text{video}}$: \emph{``Does the code generate and save a video?''}
\end{itemize}
The LLM is instructed to answer each question strictly with \textsc{True}/\textsc{False} (we parse case-insensitively and map \emph{Yes/No} to \textsc{True}/\textsc{False}). Let $z_j\in\{\textsc{True},\textsc{False}\}$ denote the LLM’s answer to the $j$-th question in $\mathcal{Q}'_x$. The reward is
\begin{equation}
r_{\mathrm{binary}}^{\mathrm{LLM}}(c\,|\,x)
\;=\;
\mathbf{1}\!\left[\forall\,j,\ z_j=\textsc{True}\right].
\label{eq:reward-pass-llm}
\end{equation}
Unlike the VLM-based variant, this setting relies solely on textual inspection of $c$; the execution and video-generation checks are text-only approximations and may disagree with actual runtime behavior. See \tabref{tab:judge_results} for an empirical comparison, and \tabref{tab:benchmark_judge} for a quantitative evaluation.

\begin{table*}[t]
\centering
\begin{tabular}{lrrrrrrrr}
\toprule
\multirow{2}{*}{\textbf{Model}} 
& \multicolumn{4}{c}{\textbf{Avg@8}} & \multicolumn{4}{c}{\textbf{Pass@8}} \\
\cmidrule(lr){2-5} \cmidrule(lr){6-9}
& \textbf{E.R.} & \textbf{R.R.} & \textbf{P.R.} & \textbf{Acc.} & \textbf{E.R.} & \textbf{R.R.} & \textbf{P.R.} & \textbf{Acc.} \\ 
\midrule
VLM-as-judge & 99.1      & 98.1  & 81.5   & 11.1   & 100.0   & 100.0   & 95.2   & 34.4  \\
LLM-as-judge & 98.6      & 94.9  & 45.4   & 7.7   & 100.0   & 100.0   & 80.2   & 29.3  \\
\bottomrule
\end{tabular}
\caption{Performance on \newtask under different reward judges. }
\label{tab:benchmark_judge}
\end{table*}

\begin{figure*}[t]
  \centering
  \includegraphics[width=.8\textwidth]{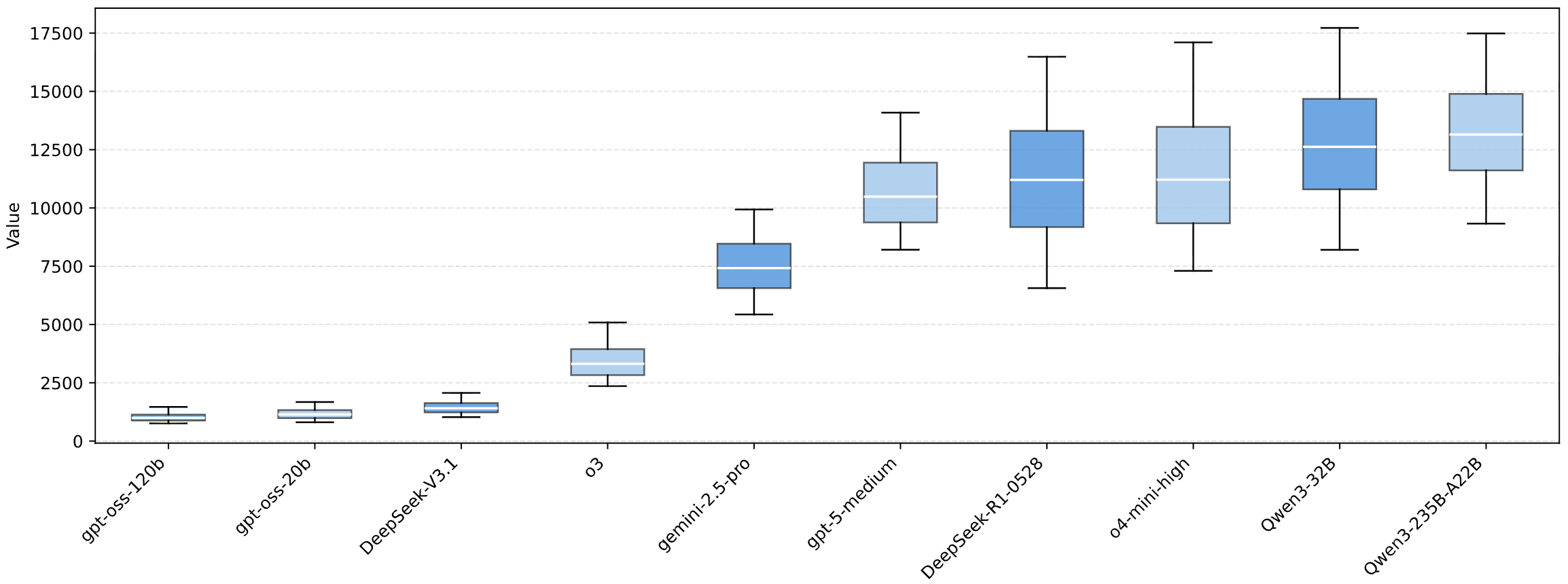}
  \caption{Distribution of response token lengths across models on the \newtask test set. The x-axis lists models, and the y-axis reports the average token counts of their generated answers.}
  \label{fig:response_token_distribution}
\end{figure*}

\begin{figure*}[t]
  \centering
  \includegraphics[width=0.75\linewidth]{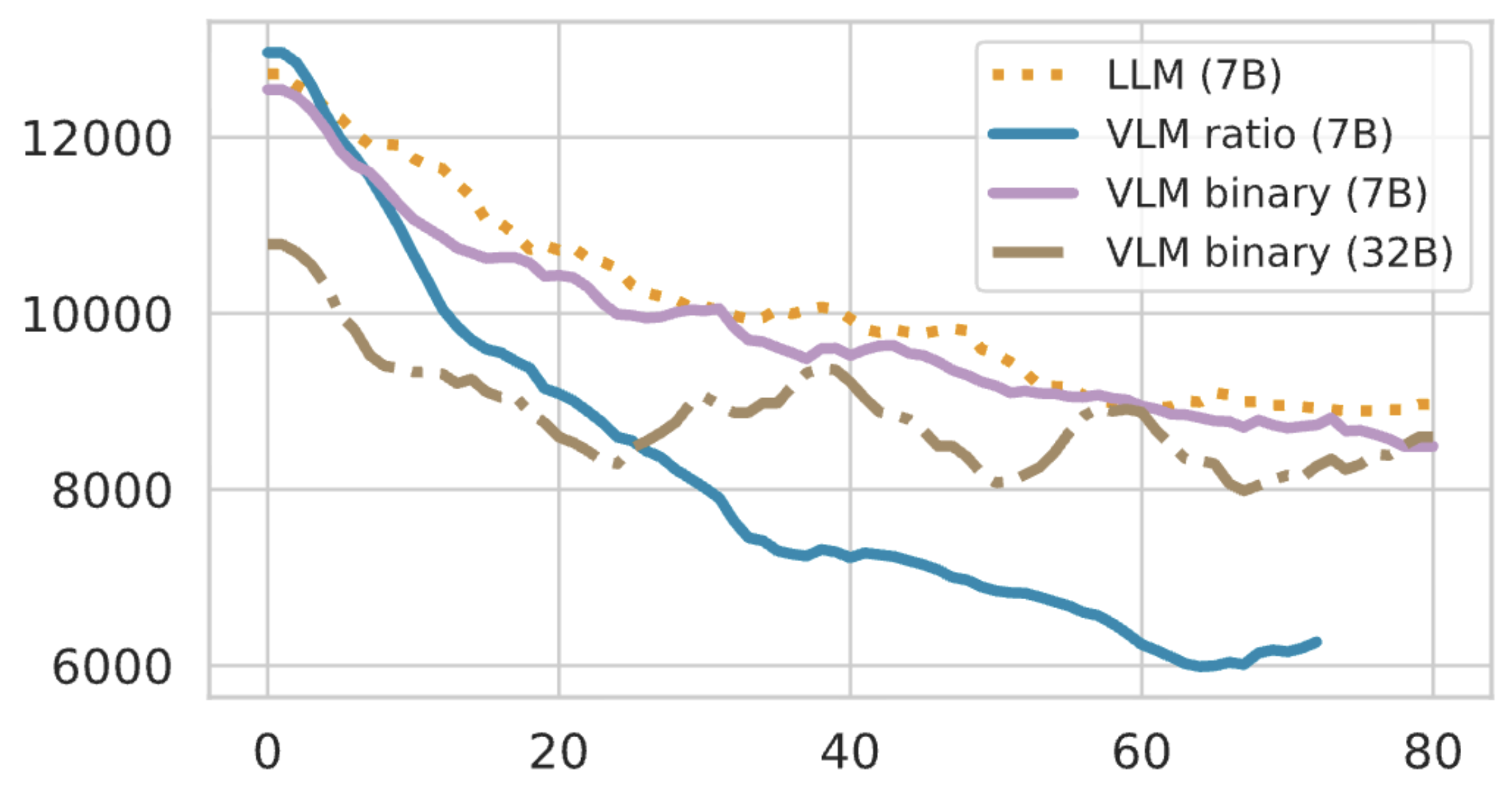}
  \caption{Response length changes during RL training.}
  \label{fig:rl_lengths}
\end{figure*}

\subsection{Computational Overhead of VLM Reward}
We further analyze the computational overhead introduced by VLM-based rewards during RL training. In our RL experiments, the \emph{LLM-as-judge} text-only reward requires an average of approximately 296 seconds per training step. In contrast, the \emph{VLM-as-judge} reward, which executes the generated code, renders the corresponding video, and performs visual judgment, requires an average of approximately 579 seconds per step, with a maximum of 1,380 seconds for more complex scenarios. The additional cost mainly comes from video rendering and visual inference.

Despite this overhead, the VLM-based reward provides substantially more faithful supervision. As shown in \tabref{tab:judge_results}, \emph{VLM-as-judge} achieves 87.8\% agreement with human annotations, compared with 69.1\% for \emph{LLM-as-judge}. This trade-off suggests that although VLM-based rewards are more computationally expensive, they better align the training signal with the end-to-end objective of generating visually aligned animations from executable code.

\subsection{Controlled Analysis of RL Gains}
To better understand where the gains from RL arise, we conduct a controlled analysis over matched subsets of generated responses. For each upstream pipeline stage, we retain only the response entries for which the Base, SFT, and RL models all successfully reach that stage. Specifically, the Same E.R., Same R.R., and Same P.R. settings respectively retain the common entries for which all three model stages produce executable code, successfully render a video, or produce a playable video. We then compare downstream performance on these shared subsets, thereby isolating improvements in visual alignment and qualitative dynamics from differences in execution, rendering, or playability.

\begin{table*}[t]
\centering
\small
\setlength{\tabcolsep}{2pt}
\begin{tabular}{llcccccccc}
\toprule
Setting & Model & Avg@8 E.R. & Avg@8 R.R. & Avg@8 P.R. & Avg@8 Acc. & Pass@8 E.R. & Pass@8 R.R. & Pass@8 P.R. & Pass@8 Acc. \\
\midrule
Same E.R. (7B) & Base    & 57.9 & 2.2  & 0.2  & 0.0  & 99.1  & 10.2  & 1.8  & 0.0  \\
Same E.R. (7B) & w/ SFT  & 57.9 & 53.2 & 19.4 & 2.8  & 99.1  & 98.5  & 62.3 & 14.1 \\
Same E.R. (7B) & w/ RL   & 57.9 & 57.3 & 47.9 & 6.8  & 99.1  & 98.8  & 92.2 & 26.9 \\
\midrule
Same R.R. (7B) & Base    & 2.1  & 2.1  & 0.2  & 0.0  & 9.3   & 9.3   & 1.8  & 0.0  \\
Same R.R. (7B) & w/ SFT  & 2.1  & 2.1  & 0.9  & 0.1  & 9.3   & 9.3   & 4.8  & 0.6  \\
Same R.R. (7B) & w/ RL   & 2.1  & 2.1  & 1.8  & 0.2  & 9.3   & 9.3   & 8.4  & 1.5  \\
\midrule
Same P.R. (7B) & Base    & 0.04 & 0.04 & 0.04 & 0.0  & 0.3   & 0.3   & 0.3  & 0.0  \\
Same P.R. (7B) & w/ SFT  & 0.04 & 0.04 & 0.04 & 0.0  & 0.3   & 0.3   & 0.3  & 0.0  \\
Same P.R. (7B) & w/ RL   & 0.04 & 0.04 & 0.04 & 0.0  & 0.3   & 0.3   & 0.3  & 0.0  \\
\midrule
Same E.R. (32B) & Base   & 97.4 & 94.2 & 49.7 & 10.8 & 100.0 & 100.0 & 82.6 & 30.5 \\
Same E.R. (32B) & w/ SFT & 97.4 & 95.0 & 74.0 & 17.4 & 100.0 & 100.0 & 91.9 & 43.1 \\
Same E.R. (32B) & w/ RL  & 97.4 & 95.2 & 88.3 & 35.9 & 100.0 & 100.0 & 98.5 & 72.2 \\
\midrule
Same R.R. (32B) & Base   & 90.8 & 90.8 & 48.0 & 10.5 & 100.0 & 100.0 & 82.3 & 29.9 \\
Same R.R. (32B) & w/ SFT & 90.8 & 90.8 & 70.7 & 16.8 & 100.0 & 100.0 & 91.3 & 42.5 \\
Same R.R. (32B) & w/ RL  & 90.8 & 90.8 & 84.6 & 34.3 & 100.0 & 100.0 & 98.2 & 71.6 \\
\midrule
Same P.R. (32B) & Base   & 36.0 & 36.0 & 36.0 & 8.3  & 71.3  & 71.3  & 71.3 & 25.1 \\
Same P.R. (32B) & w/ SFT & 36.0 & 36.0 & 36.0 & 8.8  & 71.3  & 71.3  & 71.3 & 27.2 \\
Same P.R. (32B) & w/ RL  & 36.0 & 36.0 & 36.0 & 15.7 & 71.3  & 71.3  & 71.3 & 43.1 \\
\bottomrule
\end{tabular}
\caption{Controlled analysis of RL gains under matched upstream pipeline stages. Same E.R., Same R.R., and Same P.R. denote subsets where the compared models reach the same executable, rendered, or playable stage, respectively. Results show that RL improves accuracy even when upstream execution, rendering, and playability are controlled, especially for the 32B model.}
\label{tab:rl_gain_decomposition}
\end{table*}

\tabref{tab:rl_gain_decomposition} shows the results. For the 7B model, the Same P.R. subset contains extremely few cases, as indicated by the near-zero shared playable rate in \tabref{tab:rl_gain_decomposition}, making this subset statistically unreliable. For the 32B model, however, RL consistently improves accuracy across all controlled subsets. In particular, under the Same P.R. setting, where execution, rendering, and playability are controlled, RL still improves Avg@8 accuracy from 8.8\% to 15.7\% and Pass@8 accuracy from 27.2\% to 43.1\%. These results suggest that RL improves the model's ability to generate visually aligned animations that capture the expected qualitative dynamics, rather than merely improving formatting, execution, or rendering quality.

\subsection{Response Length Analysis}
\paragraph{Response Length for Tested Models.} We further analyze the response lengths of different models on the \newtask test set, as shown in \figref{fig:response_token_distribution}. The distribution reveals a clear divide between models with and without explicit reasoning traces. Models tested with reasoning or thinking modes—including o3, DeepSeek-R1-0528, Gemini-2.5-pro, gpt-5-medium, o4-mini-high, Qwen3-32B, and Qwen3-235B-A22B—generate substantially longer responses (median $\sim$2k-15k tokens), with Qwen3-235B-A22B producing the longest outputs (extending up to $\sim$17k tokens), followed by reasoning-specialized models like o3 and DeepSeek-R1-0528 (median $\sim$12k-15k tokens). In contrast, models tested without thinking modes—DeepSeek-V3.1, gpt-oss-20b, and gpt-oss-120b—produce much more compact responses (median $\sim$1k-1.5k tokens), as they directly generate code without explicit reasoning traces.
Interestingly, this substantial token overhead ($\sim$10x difference) does not always translate to proportional performance gains. DeepSeek-R1-0528 achieves the highest Avg@8 accuracy (21.5\%) with extensive reasoning traces, while GPT-5-medium reaches competitive accuracy (20.5\%) with moderate token usage. More notably, compact models without thinking modes can still achieve reasonable performance: DeepSeek-V3.1 attains 14.5\% accuracy and gpt-oss-120b reaches 14.0\%, despite using only $\sim$1/10 of the tokens. This suggests that while chain-of-thought reasoning provides benefits for physics-inspired animation generation, efficient direct code generation remains a viable approach for resource-constrained scenarios.

\paragraph{Response Length for Trained Models.}
\figref{fig:rl_lengths} shows how response lengths change during RL training under different reward functions. As shown, the response lengths of all three models first decrease and then fluctuate as training proceeds. The models initially produce responses of around 12,000 tokens, and eventually stabilize between 6,000 and 9,000 tokens. This suggests that RL training reduces overly long reasoning traces while maintaining sufficient detail for animation-oriented code generation.

\section{Additional Experiments}
\subsection{Dataset Mixing}
Unlike existing physics reasoning benchmarks, which frame evaluation as discriminative tasks such as multiple-choice QA or numerical answer prediction, SimuScene requires models to perform a different generative task: producing executable code whose rendered animation is visually aligned with a physical scenario description and captures the expected qualitative dynamics. This places prior physics reasoning benchmarks such as SeePhys~\citep{xiang2025seephysdoesseeinghelp} in a complementary role: they evaluate physical reasoning over given inputs, whereas SimuScene evaluates generative text-to-code-to-video consistency. To contextualize our results within the broader landscape, we mix our SFT data with OpenThoughts, a general reasoning dataset covering math, science, code, and puzzles~\citep{guha2025openthoughts}, and analyze performance on the mixture. We use the full SimuScene SFT set and 10k examples from OpenThoughts. Results are shown in \tabref{tab:sft_mix}. We measure performance on math (AIME~\citep{aime}), coding (HumanEval+ and MBPP+~\citep{evalplus}), and physics reasoning benchmarks (UGPhysics~\citep{xu2025ugphysicscomprehensivebenchmarkundergraduate} and PhysReason~\citep{zhang2025physreason}).

Training exclusively on SimuScene SFT data leads to a notable decline in mathematical reasoning performance (e.g., AIME24: 46.67 → 32.24), which is expected given that our training data consists predominantly of animation-oriented code rather than mathematical reasoning traces. However, two observations mitigate this concern. First, mixing SimuScene SFT data with a general reasoning dataset (OpenThoughts, 10k examples) largely recovers mathematical performance (AIME24: 46.15) while simultaneously yielding substantial gains in coding benchmarks (HumanEval+: 39.02 → 60.37, MBPP+: 35.71 → 45.77). Second, the primary use case for SimuScene-trained models is physics-inspired animation generation, a setting where mathematical competition performance is not the target capability. We therefore recommend the mixed training recipe (Ours + OT) as the practical default, which improves animation-oriented code generation and coding performance while largely preserving mathematical reasoning ability. The remaining gap on UGPhysics and PhysReason further suggests that SimuScene targets a distinct generative capability rather than conventional physics QA.

\begin{table*}[t]
\centering
\small
\setlength{\tabcolsep}{3.2pt}
\begin{tabular}{lrrrrrr}
\toprule
\textbf{Task} & \textbf{AIME24} & \textbf{AIME25} & \textbf{HE+} & \textbf{MBPP+} & \textbf{UGPhys.} & \textbf{PhysReason} \\
\midrule
Base         & 46.67 & 35.26 & 39.02 & 35.71 & 16.68 & 43.27 \\
w/ Ours      & 32.24 & 25.05 & 52.44 & 32.28 & 11.13 & 36.17 \\
w/ OT        & 46.61 & 32.97 &  7.93 & 11.64 & 12.67 & 39.20 \\
w/ Ours+OT   & 46.15 & 33.13 & 60.37 & 45.77 & 12.73 & 38.74 \\
\bottomrule
\end{tabular}
\caption{Performance on math, code, and physical reasoning tasks when mixing SFT data with 10k OpenThoughts (OT) examples. Math tasks include AIME24 and AIME25; code tasks include HumanEval+ (HE+) and MBPP+; physics reasoning tasks include UGPhysics and PhysReason.}
\label{tab:sft_mix}
\end{table*}
\section{Additional Qualitative Examples}
\begin{figure*}[t]
  \centering
  \includegraphics[width=\textwidth]{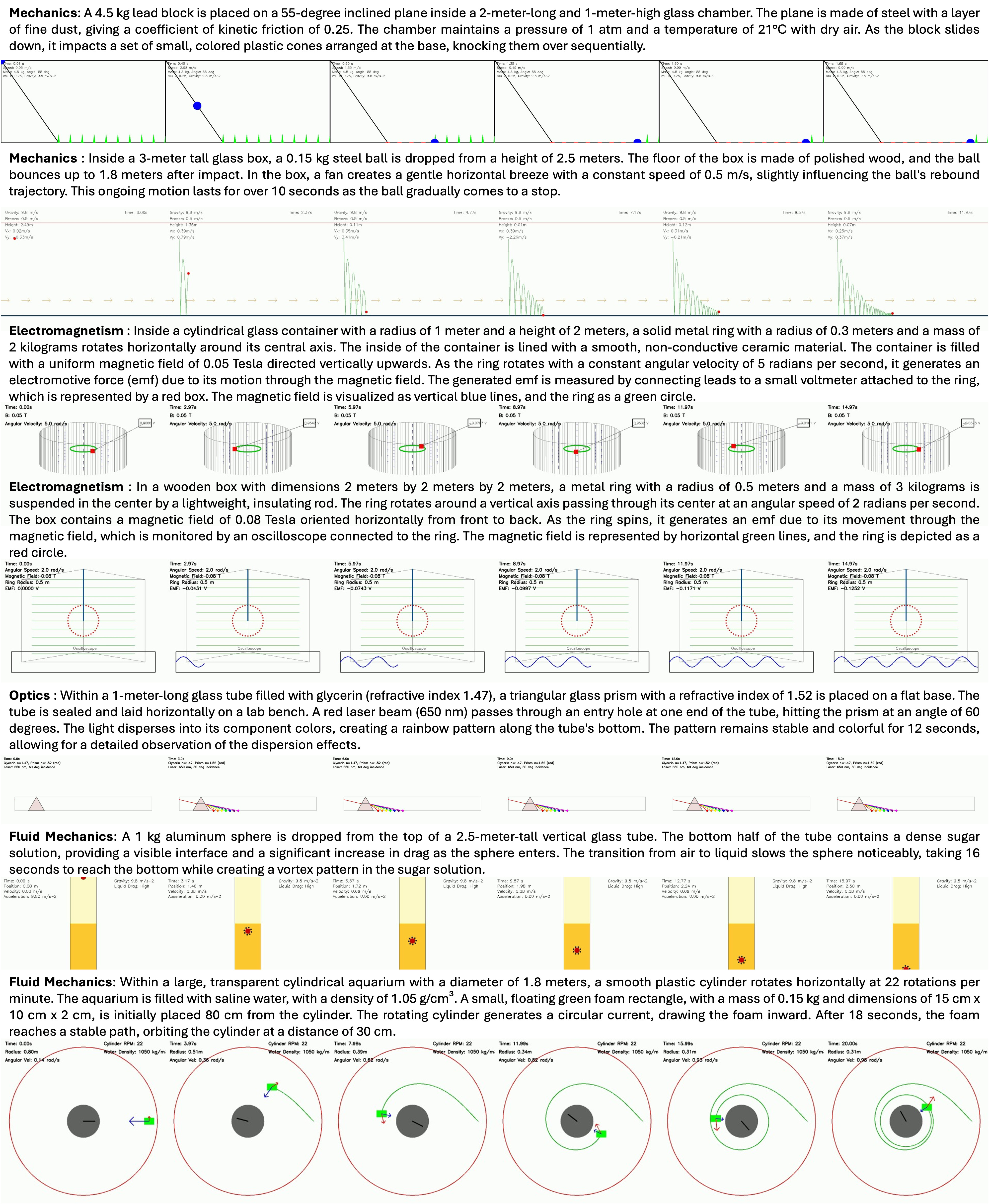}
  \caption{More representative text–to–code–to–video examples from the \newtask benchmark. Each video is rendered from LLM-generated code to produce a physics-inspired animation of the described physical scenario.}
  \label{fig:data_example2}
\end{figure*}

\figref{fig:data_example2} presents additional representative text--to--code--to--video examples, where LLM-generated programs are executed to produce physics-inspired animations from natural-language descriptions of physical scenarios. These examples highlight the diversity of dynamic scenarios covered by \newtask and demonstrate the feasibility of generating visually interpretable animations that capture qualitative motion patterns through executable code.

\begin{figure*}
  \centering
  \includegraphics[width=.9\textwidth]{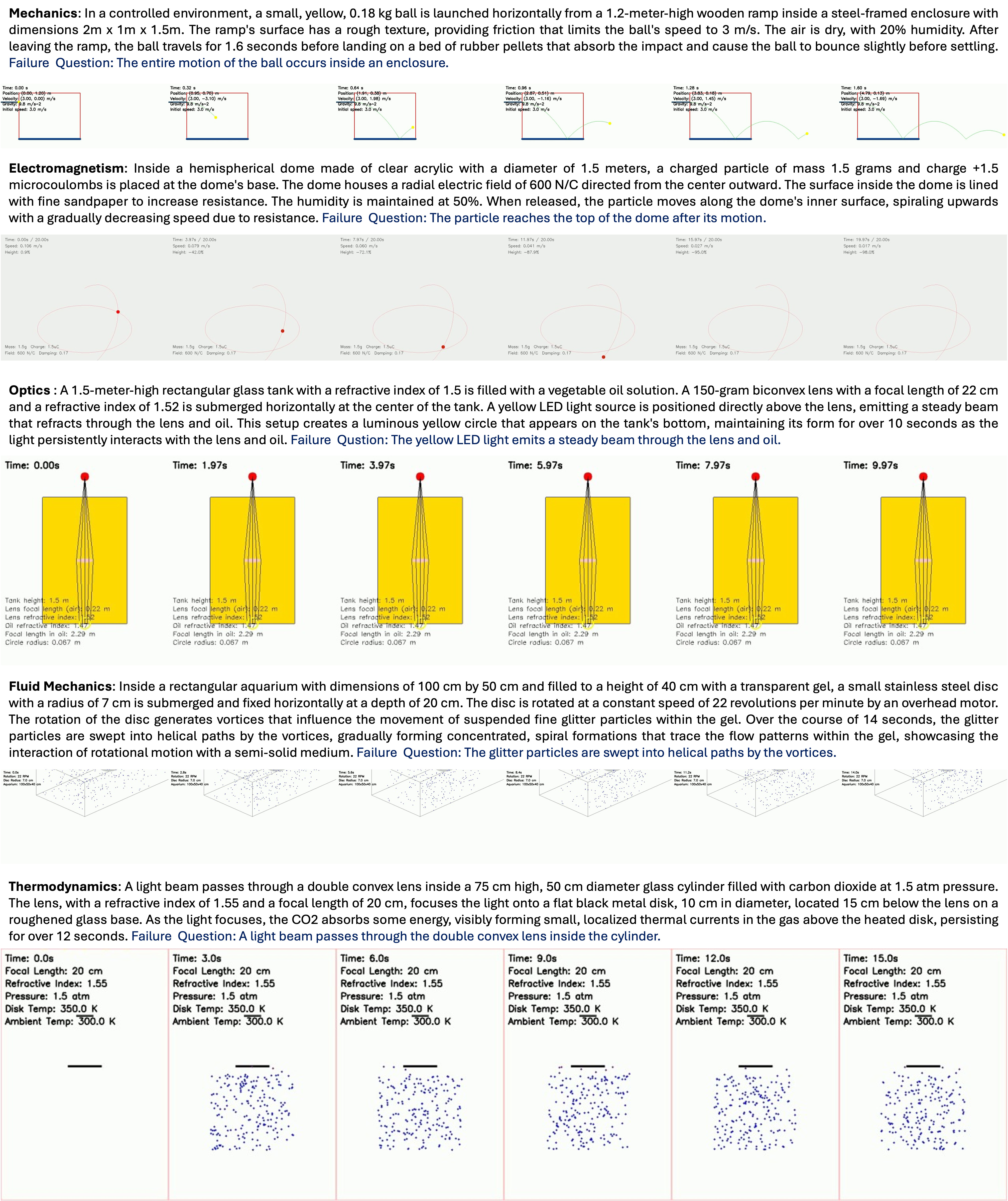}
  \caption{Representative inconsistency cases from the VLM judgment. Each example presents a text–to–code–to–video pipeline result where the VLM flagged mismatches (highlighted in blue) between the textual description and the rendered video. These failures capture common error modes, including missing or incomplete motions, object penetration through solid boundaries, and mismatches between described and rendered qualitative dynamics.}
  \label{fig:data_failure}
\end{figure*}

\figref{fig:data_failure} shows representative failure cases identified by the VLM-based judge. In these examples, the VLM identifies inconsistencies between the textual descriptions and the rendered videos, including missing or incomplete motions, object penetration through solid boundaries, and discrepancies between described and rendered qualitative dynamics. These cases demonstrate the reliability of the VLM-based judge in capturing common error modes in text--to--code--to--video pipelines, highlighting the importance of visual grounding for evaluation.

\figref{fig:data_humanvalid} presents a representative disagreement case between Qwen2.5-VL-72B-Instruct and human judgments. Despite such disagreements, the overall agreement between the VLM-based judge and human evaluation reaches 87.8\%, indicating strong consistency in most cases. This analysis helps identify the remaining limitations of current VLM models, particularly in handling nuanced physical or semantic inconsistencies, and provides insights for future improvements.

\begin{figure*}[t]
  \centering
  \includegraphics[width=\linewidth]{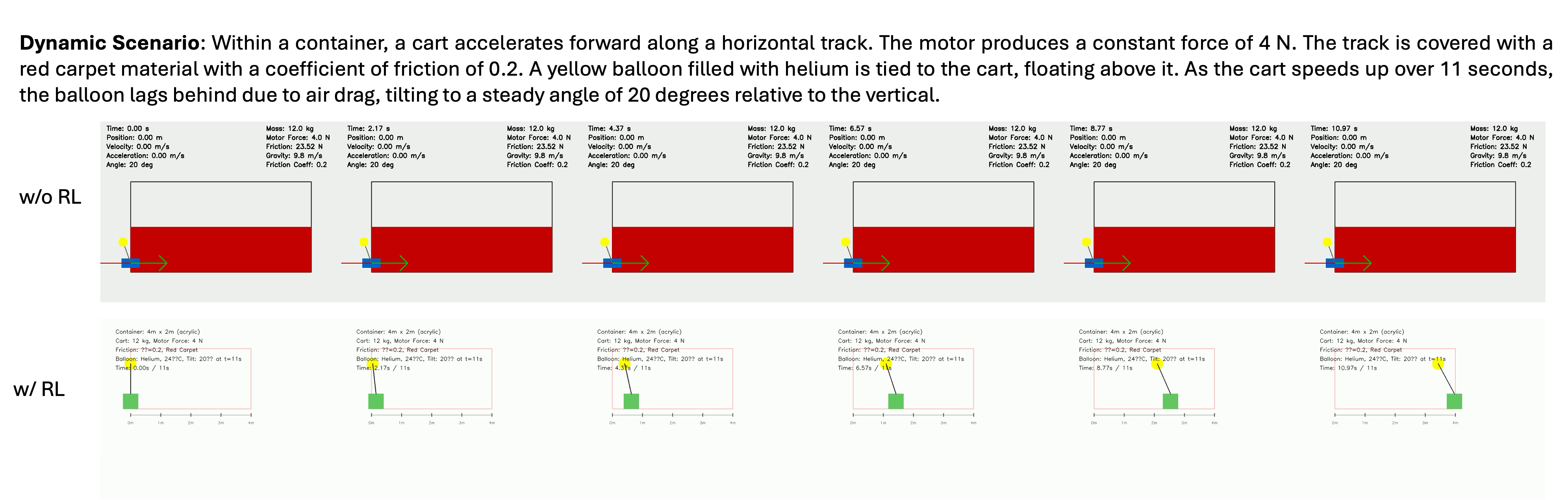} 
  \caption{A text-to-code-to-video example with and without RL.}
  \label{fig:rl_show}
\end{figure*}

Finally, \figref{fig:rl_show} provides a side-by-side comparison of model outputs before and after RL training on the same scenario. The example demonstrates how RL training improves the model's adherence to textual constraints and helps produce visually aligned animations that better capture the expected qualitative dynamics. This qualitative improvement aligns with the quantitative gains observed in our benchmark evaluations.

\section{Prompts Used in This Paper}
The following listings show the prompts used in our data construction, code generation, and evaluation pipelines. Some prompts use terms such as ``simulation'' because they were written as practical instructions for generating educational physics visualizations. In the paper, we use the term \emph{physics-inspired animation generation} to more precisely describe the evaluated task: generating executable code whose rendered animations are visually aligned with qualitative physical scenario descriptions, rather than producing numerically precise physical simulations.

We construct physical scenarios through a multi-stage generation process. Throughout the three scenario-construction stages, we use the shared system prompt in Listing~\ref{lst:system-prompt-code-generation} to establish the model as a computational physicist with expertise in scientific visualization and programming. For each physics topic, we first generate a diverse collection of basic scenarios that can be represented using simple visual elements, as shown in Listing~\ref{lst:basic-question-generation-prompt}. We separately generate additional visualizable conditions that extend these basic scenarios and increase their diversity or complexity, as shown in Listing~\ref{lst:condition-generation-prompt}. We then combine each basic scenario with a sampled number of additional conditions to form a complete dynamic scenario, using the prompt in Listing~\ref{lst:dynamic-scenario-generation-prompt}.

\begin{listing*}[t]
\begin{lstlisting}[basicstyle=\small\ttfamily,breaklines=true]
You are an expert computational physicist specializing in scientific visualization and simulation.

You are also an excellent programmer.

Your expertise includes creating educational physics simulations that effectively communicate complex physical phenomena to diverse audiences.
\end{lstlisting}
\caption{Shared system prompt used for scenario construction and code generation.}
\label{lst:system-prompt-code-generation}
\end{listing*}

\begin{listing*}[t]
\begin{lstlisting}[basicstyle=\small\ttfamily,breaklines=true]
Generate as many distinct and simple visualizable physics problems as possible related to the topic of "{topic}". Each problem should describe a clearly different type of motion or phenomenon, using basic geometric shapes, such as a ball, cube, beam, or dot, to represent objects.

Avoid repeated or similar motion types. For example, "a ball falls down" and "an apple dropped from a tree" represent the same motion and should not both be included.

Exclude scenarios involving temperature, biology, weather, microscopic physics, or effects that are not easily observable over time.

Return only a plain list of concise problem descriptions, separated by the "@" symbol.

For example:

a ball falling freely under gravity@a beam of light striking a horizontal surface@red liquid diffusing into water

Do not include duplicates or closely related variations.
\end{lstlisting}
\caption{Prompt used to generate basic visualizable physics problems for each topic.}
\label{lst:basic-question-generation-prompt}
\end{listing*}

\begin{listing*}[t]
\begin{lstlisting}[basicstyle=\small\ttfamily,breaklines=true]
Generate a diverse list of unique and realistic conditions that can be added to or used to extend the topic "{topic}" in everyday physical scenarios. Each condition must be:

- Visually clear and easy to program.
- Distinct and non-redundant.
- Physically meaningful at the human-observable scale.
- Free of temperature-related, biological, weather-related, micro-scale, or instantaneous effects that are difficult to observe or simulate.

Return the result as a plain list separated by the "@" symbol.

Do not include atomic-level effects, microscopic interactions, or abstract forces that cannot be clearly visualized.

Each condition should increase the complexity or variety of the "{topic}"-related scenario in a macroscopically observable and visualizable way.

Example:

confining the object's motion within a container@adding multiple moving objects@introducing rotational dynamics@applying time-dependent changes
\end{lstlisting}
\caption{Prompt used to generate additional visualizable conditions for each physics topic.}
\label{lst:condition-generation-prompt}
\end{listing*}

\begin{listing*}[t]
\begin{lstlisting}[basicstyle=\small\ttfamily,breaklines=true]
Generate {question_num} unique, logically consistent, and visually descriptive physics problems based on "{basicquestions[question_index]}". Use real-world, Earth-based scenarios that occur within a closed container or controlled setup. Avoid space-based or abstract environments.

Each problem must satisfy the following requirements:

1. Incorporate {cond_num + 1} distinct and non-overlapping conditions selected from {all_conditions}, with specific numeric parameter values.
2. Be described as a self-contained paragraph featuring a unique physical process that lasts more than 10 seconds.
3. Be fully visualizable using basic colored shapes and lines to represent key elements, such as a ball, inclined plane, liquid, or pulley.
4. Provide precise physical settings, including object mass, slope angle, container dimensions, surface materials, and environmental conditions.
5. Avoid motion contradictions, such as an object slowing down after release under gravity without resistance.
6. Avoid repeating motion patterns or setup logic across the generated problems.
7. Do not reference video rendering, animation, or simulation.
8. Only return the generated problems, and do not include any other information.

Return the results in the following format:

-1- First problem.
-2- Second problem.
-3- Third problem.
-4- ...
\end{lstlisting}
\caption{Prompt used to generate dynamic physics scenarios from basic problems and visualizable conditions.}
\label{lst:dynamic-scenario-generation-prompt}
\end{listing*}

After scenario generation, we use the prompt in Listing~\ref{lst:plausibility-and-verification-question-prompt} for two related purposes. It first determines whether a generated scenario is physically plausible, internally consistent, macroscopically visualizable, and feasible to represent using Python code. For each valid scenario, the same prompt generates at least five binary visual verification questions that describe the key motion trajectories and physical interactions expected to appear in the rendered video. Scenarios judged invalid are discarded, while valid scenarios and their verification questions are retained for subsequent processing. During subsequent automatic and human verification, questions may be revised, removed, or supplemented, resulting in 4--12 questions per final scenario.

\begin{listing*}[t]
\begin{lstlisting}[basicstyle=\small\ttfamily,breaklines=true]
Determine whether the question "{questions[j]}" is physically plausible, internally consistent, macroscopically visualizable, and feasible to represent using Python code.

If it is not, return only:

no

If it is plausible, generate a set of True/False verification questions in JSON format to help verify whether an input video accurately reflects the user description. The user description is:

"{questions[j]}"

Base the questions only on what is described in the user content. Ignore implementation details and avoid adding extra assumptions.

The goal is to help a human reviewer quickly check whether the video is faithful to the user's request. The questions should be comprehensive enough to determine whether the motion process follows the laws of physics, such as whether the direction of motion after a collision is logical, whether a ball moves downward during free fall, or whether the motion exceeds the range of the container.

Instructions:

1. Generate at least five clear and focused True/False questions that are comprehensive enough to determine whether the motion process follows the laws of physics.
2. Each question must directly verify a visually observable element from the description.
3. Focus on key aspects:
   - Object motion or trajectories.
   - Number of bounces or interactions.
   - Triggered events or sensors.
   - Boundary conditions.
   - Visual clarity or layout.
4. Avoid abstract, subjective, or inferred content. Avoid time-related questions, such as speed, acceleration, and distance. Do not guess details that are not in the description.
5. Do not reference Python code. Focus only on what the video is supposed to show.
6. Do not focus on unrelated appearance questions, such as whether a plane is yellow or a ball is steel.
7. Do not focus on text displayed in the video, such as whether the ball's weight matches a label or whether the video shows a length value.
8. Avoid specific numeric values, such as 70 cm, 60 degrees, or 2.5 seconds.
9. Avoid questions that are not related to the motion process, such as whether the red ball stands out against the background, whether the inclined plane surface is polished, or whether the setup is brightly illuminated.

Output format:

Return a JSON list of questions. Each item must contain:
- "id": an integer starting from 1
- "question": the verification question
- "answer": the answer to the question
- "type": "true_false"

Example:

[
  {
    "id": 1,
    "question": "The ball begins at the top of the inclined plane.",
    "answer": "True",
    "type": "true_false"
  },
  ...
]

Return only "no" or the JSON list.
\end{lstlisting}
\caption{Prompt used for plausibility judging and visual verification-question generation.}
\label{lst:plausibility-and-verification-question-prompt}
\end{listing*}
We further assess whether the generated verification questions provide sufficient coverage of the complete motion described by the scenario. The sufficiency-judging prompt in Listing~\ref{lst:sufficiency-judge-prompt} assumes that all verification questions are answered affirmatively and determines whether this would be sufficient to establish that the entire described process is correctly represented. If a question set is judged insufficient, it is routed back to the verification-question generation stage for revision or regeneration. The prompt also asks for optional question weights as part of its original output schema; however, these weights are not used in our final data filtering, evaluation, or training pipeline.

\begin{listing*}[t]
\begin{lstlisting}[basicstyle=\small\ttfamily,breaklines=true]
You are an annotator.

You are given a motion description:

{question}

You are also given a set of related visual verification questions:

{all_vlm_questions}

Assume that the answers to all of these questions are "Yes".

Determine whether the questions in {all_vlm_questions} are sufficient to fully judge the entire motion described in {question}. Your response should be either "Yes" or "No".

If your answer is "Yes", summarize the key noun phrases in the motion description and describe their relationships. For example, "a red ball is moving toward a blue ball".

The output should be in JSON format with the following structure:

{
  "enableAnnotator": "Yes or No",
  "summarize": "A red ball is moving toward a blue ball.",
  "vlm_questions": [
    {
    "index": "index number",
    "question": "question1",
    "weight": "0-1 scale, where 0 is not important and 1 is very important"
    }
  ]
}
\end{lstlisting}
\caption{Prompt used for sufficiency judging of visual verification questions.}
\label{lst:sufficiency-judge-prompt}
\end{listing*}
To evaluate the capabilities of existing LLMs, we reuse the shared system prompt in Listing~\ref{lst:system-prompt-code-generation} together with the detailed code-generation user prompt in Listing~\ref{lst:user-prompt-code-generation}. While the shared system prompt provides a consistent expert role across both scenario construction and code generation, the user prompt specifies the implementation language, rendering library, video duration, frame rate, output format, visual presentation, and expected code structure. We use this relatively comprehensive prompt to provide off-the-shelf models with explicit task guidance and to elicit their strongest possible performance on physics-inspired animation generation. This detailed evaluation prompt is distinct from the simpler and more diverse instructions used during model training.

\begin{listing*}[t]
\begin{lstlisting}[basicstyle=\small\ttfamily,breaklines=true]
Your task is to write a Python script that generates an educational video simulating a physical process. The video will be used in academic settings to help students better understand and visualize physics concepts.

Given a textual description of a physical process:

"{content}"

Write a Python script that simulates this process and outputs a video saved as:

"name.mp4"

Here, "name" is a user-defined parameter passed when running the script.

Requirements for the video output:

1. Use clear and distinct colors to represent different objects, trajectories, or forces.
2. Overlay a real-time timestamp that updates continuously throughout the simulation.
3. Display all relevant parameter values, such as gravity, speed, and angle, clearly on the screen.
4. Ensure the camera view is wide enough to fully capture the entire motion, adjusting dynamically if needed. Use the best view for the simulation so that the viewer can see the whole process.
5. Provide smooth and continuous animation at a consistent frame rate of 30 FPS.
6. Maintain a clean, uncluttered visual style with minimal distractions and a neutral background.
7. Keep the video duration between 10 and 20 seconds, slow enough to allow viewers to observe and understand the key transitions.
8. Save the output as an MP4 video in a suitable resolution of at least 360p.
9. When the process finishes, the video should also finish.
10. Use OpenCV to generate the video, and ensure that the code is correct, complete, and runnable without errors.

Focus on clarity, interpretability, and visual appeal to make the video intuitive and easy to understand for both technical and non-technical audiences.

Physics simulation:

- Implement precise physical equations.
- Use appropriate time steps for smooth motion.
- Include relevant force vectors and trajectories.

The final code should:

1. Initialize all necessary libraries and variables.
2. Set up the video writer with specified parameters.
3. Implement the physics calculations.
4. Create and save the animation.
5. Include error handling and resource cleanup.

Ensure that the code follows PEP 8 style guidelines and includes comments explaining key components. The simulation should prioritize educational value while maintaining scientific accuracy.

Instructions:

1. Output only the complete Python code. Do not include explanations outside the code. For example, the code may start with:

import cv2
import numpy as np

2. The code should install any dependencies within the script if needed.
3. The code should run without errors.
4. The code should be complete and self-contained.
5. The code should be correct and physically accurate.
6. The code should be efficient and optimized.
7. The code should be easy to understand and modify.
\end{lstlisting}
\caption{User prompt for code generation.}
\label{lst:user-prompt-code-generation}
\end{listing*}

For evaluation, the generated code is executed to produce a video, which is then assessed using the VLM-as-judge prompt in Listing~\ref{lst:vlm-as-judge-prompt}. Given the rendered video and its associated verification questions, the VLM independently determines whether each statement is visually supported by the observed motion and interactions. Statements that are contradicted by the video or cannot be confirmed from the available visual evidence are judged false. A generated response passes the end-to-end evaluation only if its code successfully produces a playable video and all associated verification questions are judged true.

\begin{listing*}[t]
\begin{lstlisting}[basicstyle=\small\ttfamily,breaklines=true]
You are given a video and several visual verification questions. Your task is to judge each question as true or false based only on what can be seen or reasonably inferred from the video. If the visual evidence is insufficient to confirm the statement, or if the statement directly contradicts the video, answer "false".

Visual reasoning guidelines:

1. Perspective changes: Objects can look different from different angles. For example, a circular path may look curved, straight, or wavy depending on the viewpoint, and a cylinder may look like a circle from the top view or a rectangle from the side view.

2. Simplified or symbolic drawings: The video may use simple shapes or colors to represent real objects. If the motion or layout still matches the description, treat it as true.

3. Container boundaries: If no container is drawn, assume that the video frame is the boundary. If a container is shown, treat it as see-through only if the interior is visible. If something is not visible, do not assume that it is inside the container.

4. Shape and motion: Focus on shape and motion, and ignore assumptions about material, weight, texture, or color. If the described motion does not match the video, such as "slides down" versus actually moving up, answer false.

5. Occlusion: If something is partly hidden, use nearby visual cues to infer how it is moving or positioned.

Input:

- Questions: {questions}

Output format:

Return a JSON list of objects. Do not include conversational filler or Markdown formatting tags. Each object should contain the following fields:

- "index": The question index.
- "question": The full question text.
- "analysis": A concise step-by-step breakdown of what was observed versus what the question claims.
- "result": "true" if visually confirmed, and "false" if contradicted or if the evidence is insufficient.

Example format:

[
  {
    "index": "1",
    "question": "The ball rolls along the circular path.",
    "analysis": "The red sphere moves in a consistent arc that completes a 360-degree loop.",
    "result": "true"
  }
]
\end{lstlisting}
\caption{VLM-as-judge prompt used for visual verification.}
\label{lst:vlm-as-judge-prompt}
\end{listing*}

During model training, we do not use the fixed and detailed evaluation prompt. Instead, we generate a diverse collection of less prescriptive instruction templates to reduce dependence on a particular prompt structure and improve instruction-following generalization. We use GPT-5\citep{openai_gpt5_medium} to propose plausible personas and application contexts in which users may request Python code for visualizing physical phenomena. We then randomly sample template components, including the persona, narrative perspective, position, and formatting of the physical-scenario description, primary action verb, and additional implementation constraints.

The meta-prompt in Listing~\ref{lst:instruction-template-diversification-prompt} instructs GPT-5 to combine the sampled components into a cohesive and natural instruction template containing the literal placeholder \texttt{\{content\}}. The placeholder is subsequently replaced with a physical scenario description to form the final training instruction. Although these training prompts vary in wording, structure, and level of detail, they preserve the same core task of requesting Python code that visually represents the described physical process. For brevity, we show one complete example and omit the remaining examples in the displayed meta-prompt; the full prompt is included in the released code.
\begin{listing*}[t]
\begin{lstlisting}[basicstyle=\small\ttfamily,breaklines=true]
Task: You are a prompt engineering expert. Your goal is to generate a single, natural-sounding instruction prompt that will be used to fine-tune a smaller LLM.

Target task: The final prompt you generate must ask an AI to write Python code to simulate a physical phenomenon or motion based on a provided description.

Variables to incorporate:
1. Persona: The persona that you will use in the prompt. If the value is None, do not use a persona.
2. Identity: If the value is "You", address the AI as the persona. If the value is "Me", write the prompt from the perspective of that persona.
3. Position: Place the placeholder {content} at the beginning, middle, or end of the generated prompt.
4. Verb: Use this as the primary action word for the coding task.
5. Format: Wrap the {content} placeholder according to the formatting rules below.
6. Conditions: Integrate these technical requirements or reminders naturally into the request.

Formatting rules for {content}:
- paragraph: Use {content} as a single paragraph.
- json: Use a JSON string such as {"physical_description": "{content}"}.
- markdown: Use a Markdown code block containing {content}.
- python: Use a Python code block such as description = "{content}".

Requirements:
- The generated prompt must be a cohesive, realistic request from a user to an AI.
- The generated prompt must include the literal string {content}. This is the placeholder where the physical description will be inserted later.
- The generated prompt must clearly state that the goal is to simulate the physical phenomenon using Python.

Output: Provide only the text of the generated prompt template.

Example templates for reference:

Example 1

- Persona: High School Physics Teacher
- Identity: Me
- Position: Middle
- Verb: create
- Format: python
- Conditions: ["Recommend using opencv library."]

Generated prompt:

"I am a high school physics teacher looking for a way to show my students how forces interact. I have defined the parameters here:

description = "{content}"

Please create a Python simulation based on that variable. I recommend using the opencv library to draw the frame-by-frame motion of the objects so the students can see the trajectory clearly."

Example 2
...

Now, generate a new, unique prompt template using the following values:

- Persona: Robotics Developer
- Identity: Me
- Position: End
- Verb: provide
- Format: paragraph
- Conditions: ["Remind to save the video file."]
\end{lstlisting}
\caption{Meta-prompt used to generate diversified instruction templates for model training.}
\label{lst:instruction-template-diversification-prompt}
\end{listing*}

\section{Use of LLMs in This Work}

Large language models and vision-language models are used as research components in several stages of this work, including scenario and verification-question generation, verification-question sufficiency assessment, code generation, rendered-video evaluation, instruction-template diversification, and reinforcement learning with vision-based rewards. The specific models, prompts, and procedures used in these stages are described in the corresponding sections of the paper and supplementary material.

LLMs were also used as writing assistants to improve grammar, clarity, and presentation during manuscript preparation. The authors independently designed the research methodology, implemented and conducted the experiments, analyzed the results, and determined the conclusions. All model-generated content used in the research or manuscript was reviewed by the authors, who assume full responsibility for the accuracy, validity, and integrity of the work.

\section{Limitations and Future Work}

\newtask evaluates whether generated programs produce animations that are executable, visually interpretable, and qualitatively aligned with natural-language physical scenarios. It does not directly evaluate numerical simulation accuracy, calibrated physical parameters, or exact trajectories. This task boundary is intentional and supports applications such as educational visualization and rapid animation prototyping, but it also limits the conclusions that can be drawn about a model's ability to perform precise physical simulation. In addition, the current scenarios primarily use simplified visual representations and have limited temporal and interaction complexity. Future work could extend the benchmark to richer 3D environments, more complex materials and lighting, longer interaction sequences, and scenarios grounded in sensors or calibrated simulators.

A second limitation is the reliance on synthetically generated scenarios and verification questions. Although we encourage diversity through a taxonomy of 52 concepts across five physics domains and use an independent model to check question sufficiency and trigger regeneration, synthetic data may still inherit stylistic patterns, omissions, or biases from the generating models. Future versions could incorporate expert-authored scenarios, educational resources, real simulation specifications, and broader human validation. Such additions would also make it possible to study how well models generalize across synthetic, expert-written, and naturally occurring instructions.

Finally, our evaluation and reinforcement-learning pipelines rely on VLM judgments of rendered videos. VLMs may miss subtle temporal, semantic, or physical inconsistencies, and learned policies may in principle exploit systematic weaknesses in the judge. Human evaluation shows substantial agreement with the selected VLM and does not reveal systematic reward exploitation in our experiments, but these checks cannot eliminate all evaluation errors. Future work could strengthen reliability through multi-VLM adjudication, uncertainty calibration, targeted human review, temporal video models, and auxiliary signals from physics engines or programmatic trajectory checks.

\section{Reproducibility Statement}
To support the reproducibility of our work, the main paper provides detailed descriptions of the \textbf{dataset construction}, \textbf{evaluation pipeline}, and \textbf{model training} procedures for \newtask, including scenario generation, code generation, VLM-based verification, model configurations, and training hyperparameters. Additional implementation details, ablation studies, and the prompts used in our pipeline are documented in this supplementary material. For long prompts whose repetitive examples are abbreviated in the listings, the complete versions are included in the released code. We also include the code and scripts required for data construction, video rendering, evaluation, and model training in the supplementary materials. Together, these resources are intended to enable the community to reproduce our experiments and extend \newtask in future research.

\end{document}